\documentclass{article}

\usepackage[final,nonatbib]{neurips_2023}
\usepackage[numbers]{natbib}

\usepackage{xcolor}
\usepackage{amsmath,amsfonts,bm}
\usepackage[utf8]{inputenc} 
\usepackage[T1]{fontenc}    
\usepackage[hidelinks, colorlinks=true, citecolor=blue]{hyperref}       
\usepackage{url}            
\usepackage{amsfonts}       
\usepackage{nicefrac}       
\usepackage{graphicx}
\usepackage{placeins}
\usepackage{subfigure}
\usepackage{dsfont}
\usepackage{xspace}
\usepackage{amssymb}
\usepackage{mathtools}
\usepackage{hyperref}
\usepackage{amsfonts}
\usepackage{color}

\usepackage{float}
\usepackage{scalerel,stackengine}
\usepackage{soul}
\usepackage[algo2e,ruled,vlined]{algorithm2e}
\usepackage{appendix}
\usepackage{amsthm}

\usepackage{wasysym}
\usepackage{pifont}
\usepackage{bbm}
\usepackage{lipsum}

\def\eqref#1{equation~\ref{#1}}

\def\1{\bm{1}}

\def\vu{{\bm{u}}}
\def\vv{{\bm{v}}}

\DeclareMathAlphabet{\mathsfit}{\encodingdefault}{\sfdefault}{m}{sl}
\SetMathAlphabet{\mathsfit}{bold}{\encodingdefault}{\sfdefault}{bx}{n}

\newcommand{\R}{\mathbb{R}}

\usepackage{custom_edit}

\usepackage[utf8]{inputenc} 
\usepackage[T1]{fontenc}    
\usepackage[hidelinks]{hyperref}       
\usepackage{url}            
\usepackage{booktabs}       
\usepackage{amsfonts}       
\usepackage{nicefrac}       
\usepackage{microtype}      
\usepackage{tikz}
\usepackage{algorithm,algpseudocode}  
\usepackage{caption}

\usetikzlibrary{positioning}

\usepackage{xspace}

\let\olditemize=\itemize
\let\endolditemize=\enditemize
\renewenvironment{itemize}{
    \olditemize
    \setlength{\itemsep}{0pt}
    \setlength{\parskip}{0pt}
}{\endolditemize}

\let\oldenumerate=\enumerate
\let\endoldenumerate=\endenumerate
\renewenvironment{enumerate}{
    \oldenumerate
    \setlength{\itemsep}{0pt}
    \setlength{\parskip}{0pt}
}{\endoldenumerate}

\providecommand{\red}{}
\renewcommand{\red}[1]{\textcolor{red}{#1}}

\renewcommand{\include}[1]{\hfill\red{DO NOT USE \texttt{include}; USE \texttt{input} INSTEAD!!!}\hfill}

\newif\iffeisha

\title{Debias Coarsely, Sample Conditionally:\\ Statistical Downscaling through Optimal Transport and Probabilistic Diffusion Models}

\author{Zhong Yi Wan\thanks{Equal contribution}\\
Google Research\\
Mountain View, CA 94043, USA \\
\texttt{wanzy@google.com} \\
\And
Ricardo Baptista$^*$\\
California Institute of Technology\\
Pasadena, CA 91106, USA \\
\texttt{rsb@caltech.edu} \\
\AND
Yi-fan Chen\\
Google Research\\
Mountain View, CA 94043, USA \\
\texttt{yifanchen@google.com}  \\
\And
John Roberts Anderson\\
Google Research\\
Mountain View, CA 94043, USA  \\
\texttt{janders@google.com} \\
\AND
Anudhyan Boral\\
Google Research\\
Mountain View, CA 94043, USA \\
\texttt{anudhyan@google.com}  \\
\And
Fei Sha\\
Google Research\\
Mountain View, CA 94043, USA  \\
\texttt{fsha@google.com} \\
\And
Leonardo Zepeda-N\'u\~nez\\
Google Research\\
Mountain View, CA 94043, USA \\
\texttt{lzepedanunez@google.com} \\
}

\begin{document}

\maketitle

\begin{abstract}
We introduce a two-stage probabilistic framework for statistical downscaling \emph{using unpaired data}. Statistical downscaling seeks a probabilistic map to transform low-resolution data from a \emph{biased} coarse-grained numerical scheme to high-resolution data that is consistent with a high-fidelity scheme. Our framework tackles the problem by
composing two transformations: (i) a debiasing step via an optimal transport map, and (ii) an upsampling step achieved by a probabilistic diffusion model with \textit{a posteriori} conditional sampling. This approach characterizes a conditional distribution \emph{without needing paired data}, and faithfully recovers relevant physical statistics from biased samples. We demonstrate the utility of the proposed approach on one- and two-dimensional fluid flow problems, which are representative of the core difficulties present in numerical simulations of weather and climate. Our method produces realistic high-resolution outputs from low-resolution inputs, by upsampling resolutions of $8\times$ and $16\times$. Moreover, our procedure correctly matches the statistics of physical quantities, even when the low-frequency content of the inputs and outputs do not match, a crucial but difficult-to-satisfy assumption needed by current state-of-the-art alternatives. Code for this work is available at: {\footnotesize\url{https://github.com/google-research/swirl-dynamics/tree/main/swirl_dynamics/projects/probabilistic_diffusion}}.
\end{abstract}

\feishatrue  

\section{Introduction}
\label{sIntro}

Statistical downscaling is crucial to understanding and correlating simulations of complex dynamical systems at multiple resolutions. For example, in climate modeling, the computational complexity of general circulation models (GCMs)~\cite{Balaji2022:GCMs} grows rapidly with resolution. This severely limits the resolution of long-running climate simulations. Consequently, accurate predictions (as in forecasting localized, regional and short-term weather conditions) need to be \emph{downscaled} from coarser lower-resolution models' outputs.  This is a challenging task: coarser models do not resolve small-scale dynamics, thus creating bias~\citep{Christensen2008,schneider2017climate,Zelinka2020}. They also lack the necessary physical details (for instance, regional weather depends heavily on local topography) to be of practical use for regional or local climate impact studies~\citep{Grotch1991:GCMforclimate,Hall2014:Projecting_regional_change}, such as the prediction or risk assessment of extreme flooding \cite{Gutmann2014:statistical_downscaling_water_resources,Hwang2014:Statistical_Downscaling_precipitations}, heat waves~\cite{Naveena2022:downscaling_temperature}, or wildfires \cite{Abatzoglou2012:downscaling_wildfire}.

At the most abstract level, \emph{statistical downscaling} \cite{Wilby_1998:downscaling,Wilby:2006} learns a map from low- to high-resolution data. However, it has several unique challenges. First, unlike supervised machine learning (ML), there is \emph{no natural pairing of samples} from the low-resolution model (such as climate models \cite{Danabasoglu2020:cesm}) with samples from higher-resolution ones (such as weather models that assimilate observations \cite{hersbach2020era5}). Even in simplified cases of idealized fluids problems, one cannot naively align the simulations in time, due to the chaotic behavior of the models: two simulations with very close initial conditions will diverge rapidly. Several recent studies in climate sciences have relied on synthetically generated paired datasets. The synthesis process, however, requires accessing both low- and high-resolution models and either (re)running costly high-resolution models while respecting the physical quantities in the low-resolution simulations~\citep{dixon2016evaluating,Huang2020:forced_paired_data} or (re)running low-resolution models with additional terms nudging the outputs towards high-resolution trajectories~\citep{charalampopoulos2023statistics}. In short, requiring data in correspondence for training severely limits the potential applicability of supervised ML methodologies in practice, despite their promising results~\citep{hammoud2022cdanet,harris2022generative,pan2021learning,price2022increasing,harder2022generating}.

\begin{figure}[t]
  \vspace{-1cm}
  \centering
  {\setlength\tabcolsep{1.5pt}
      \begin{tabular}{M{.4\linewidth}M{.55\linewidth}}
        \includegraphics[height=4.5cm, trim={0.1cm 13cm 15cm 0cm}, clip]{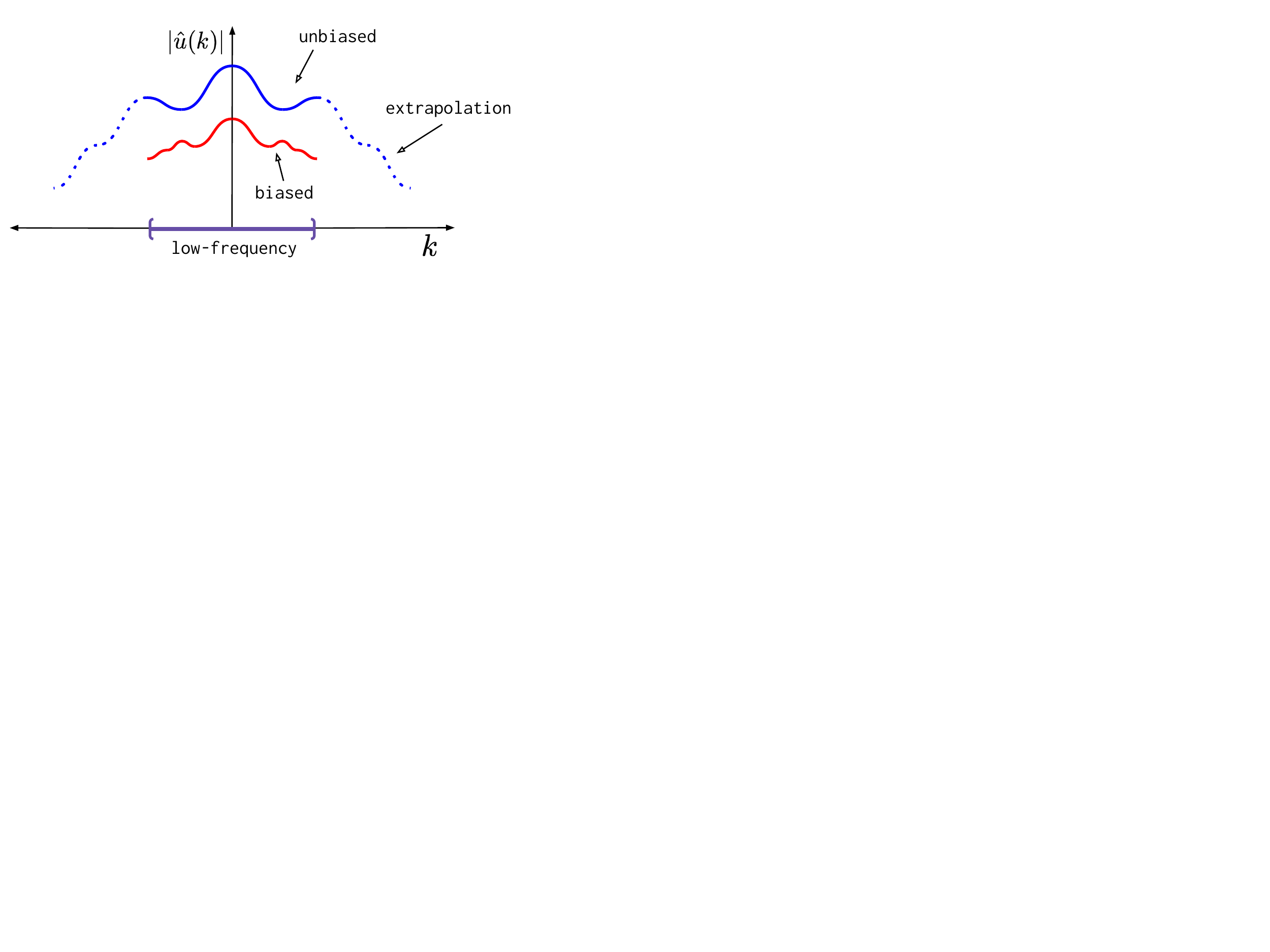}  \vspace{-25pt} &
        \hspace{2cm} \includegraphics[height=4.5cm, trim={0.1cm 4cm 8cm 0cm}, clip]{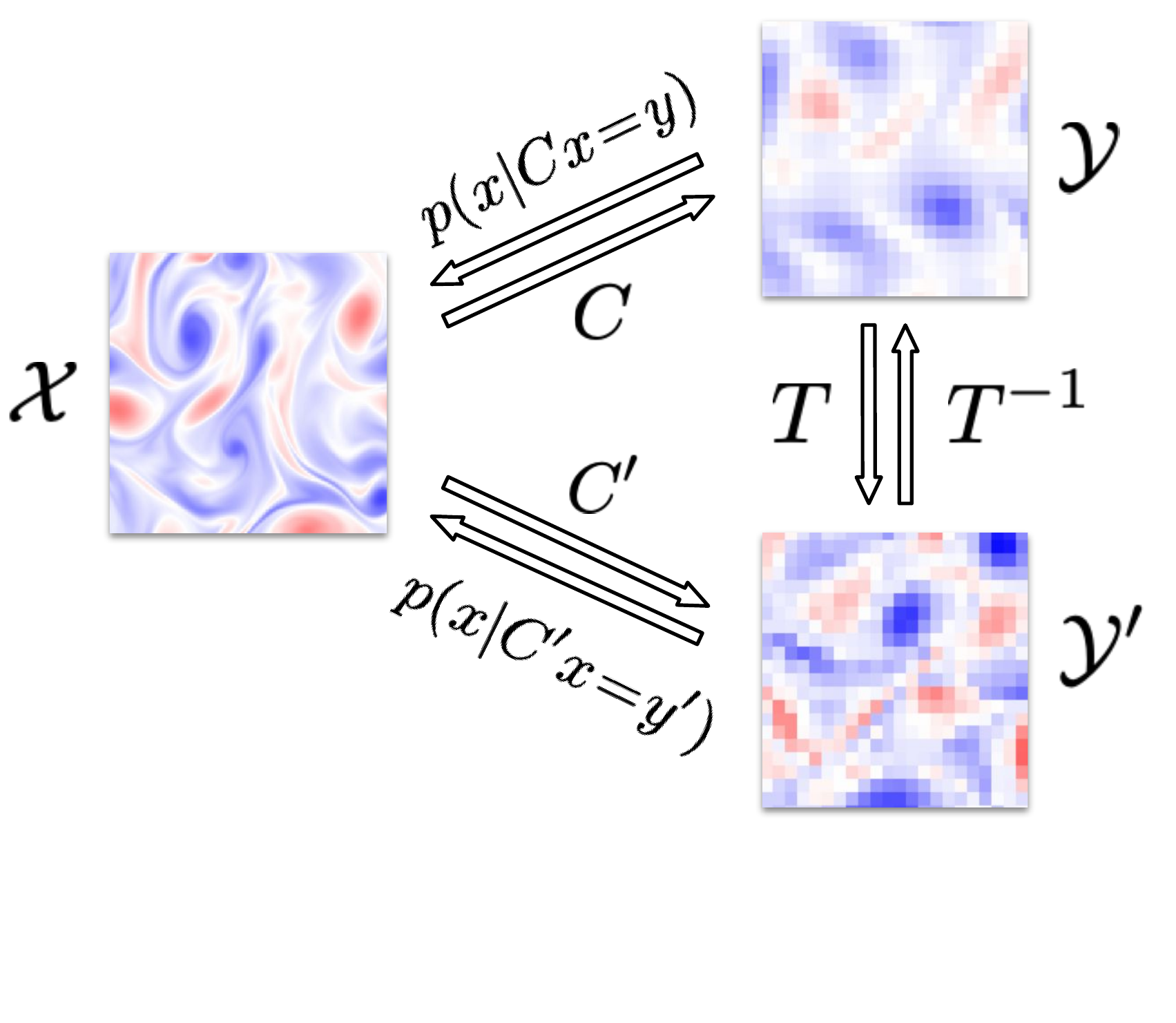} \vspace{-25pt} \\
        \hspace{30pt} (a)  & \hspace{40pt} (b) 
      \end{tabular}
  }
  \caption{(a) Upsampling (super-resolution) as frequency extrapolation in the Fourier domain. The model extrapolates low-frequency content to higher-frequencies (dashed blue). The debiasing map corrects the biased low-frequency content (solid red). (b) Illustration of the proposed framework where $\mathcal{X}$ is the space of high-resolution data, $\mathcal{Y}$ is the space of low-resolution data, $C$ is an \emph{unknown nonlinear} map linking $\mathcal{X}$ and $\mathcal{Y}$, $C'$ is a \emph{known linear} downsampling map, $\mathcal{Y}'$ is an intermediate (low-resolution) space induced by the image of $C'$, and $T$ is an invertible debiasing map such that $C$ can be factorized as $T^{-1} \circ C'$. The conditional probabilities $p(x|C'x \hspace{-2pt}=\hspace{-2pt} y')$ are used for the probabilistic upsampling procedure.}
  \label{fig:framework_diagram}
  \vspace{-.6cm}
\end{figure}

Second, unlike the setting of (image) super-resolution~\cite{dong2014learning}, in which an ML model learns the (pseudo) inverse of a downsampling operator \cite{ResLap,Vandal_2017:DeepDS}, downscaling additionally needs to correct the bias.  
This difference is depicted in Fig.~\ref{fig:framework_diagram}(a). Super-resolution can be recast as frequency extrapolation \cite{Candes2016:Superresolution}, in which the model reconstructs high-frequency contents, 
while matching the low-frequency contents 
of a low-resolution input. However, the restriction of the target high-resolution data 
may not match the distribution of the low-resolution data in Fourier space~\cite{kolmogorov1962refinement}. 
Therefore, debiasing is necessary to correct the Fourier spectrum of the low-resolution input to render it admissible for the target distribution (moving solid red to solid blue lines with the dashed blue extrapolation in Fig.~\ref{fig:framework_diagram}). Debiasing allows us to address the crucial yet challenging prerequisite of aligning the low-frequency statistics between the low- and high-resolution datasets.

Given these two difficulties, statistical downscaling should be more naturally framed as matching two probability distributions linked by an unknown map; such a map emerges from both distributions representing the same underlying physical system, albeit with different characterizations of the system's statistics at multiple spatial and temporal resolutions. The core challenge is then: \emph{how do we structure the downscaling map so that the (probabilistic) matching can effectively remediate the bias introduced by the coarser, i.e., the low-resolution, data distribution?}

Thus, the main idea behind our work is to introduce a debiasing step so that the debiased (yet, still coarser) distribution is closer to the target distribution of the high-resolution data. This step results in an intermediate representation for the data that preserves the correct statistics needed in the follow-up step of upsampling to yield the high-resolution distribution. In contrast to recent works on distribution matching for unpaired image-to-image translation~\cite{Zhu2017:CycleGAN} and climate modeling~\cite{climalign:2021}, the additional structure our work imposes on learning the mapping prevents the bias in the low-resolution data from polluting the upsampling step. We review those approaches in \S\ref{sRelated} and compare to them in \S\ref{sec:numerical_experiments}.

Concretely, we propose a new \text{probabilistic} formulation for the downscaling problem that handles \textit{unpaired data} directly, based on a factorization of the unknown map linking both low- and high-resolution distributions. This factorization is depicted in Fig.~\ref{fig:framework_diagram}(b). By appropriately restricting the maps in the factorization, we rewrite the downscaling map as the composition of two procedures: a debiasing step performed using an optimal transport map~\cite{cuturi2013sinkhorn}, which \emph{couples the data distributions} and corrects the
biases of the low-resolution snapshots; followed by an upsampling step performed using conditional probabilistic diffusion models, which have produced state-of-the-art results for image synthesis and flow construction~\cite{bischoff2022:unpaired,Li2022:SRDiff,shu2023physics,song2020score}.

We showcase the performance of our framework on idealized fluids problems that exhibit the same core difficulty present in atmospheric flows. We show that our framework is able to generate realistic snapshots that are faithful to the physical statistics, while outperforming several baselines. 

\iffeisha
\section{Related work}
\label{sRelated}




The most direct approach to upsampling low-resolution data is to learn a low- to high-resolution mapping via paired data when it is possible to collect such data. For complex dynamical systems, several methods carefully manipulate high- and low-resolution models, either by nudging or by enforcing boundary conditions, to produce paired data without introducing spectral biases
~\cite{charalampopoulos2023statistics,dixon2016evaluating}. Alternatively, if one has strong prior knowledge about the process of downsampling, optimization methods can solve an inverse problem to directly estimate the high-resolution data, leveraging prior assumptions such as sparsity in compressive sensing~\cite{Candes:2013,Candes2016:Superresolution} or translation invariance~\cite{Hua_Sarkar:1991}. 
 
In our setting, there is no straightforward way to obtain paired data due to the nature of the problem (i.e., turbulent flows, with characteristically different statistics across a large span of spatio-temporal scales). In the weather and climate literature (see \cite{vandal2017intercomparison} for an extensive overview), prior knowledge can be exploited to downscale  specific variables~\cite{Wilby_1998:downscaling}. One of the most predominant methods of this type is bias-correction spatial disaggregation (BCSD), which combines traditional spline interpolation with a quantile matching bias correction \cite{Maraun2013:quantile_matching}, and linear models~\cite{Hessami_2008:linear_models_statistical_dowscaling}.
Recently, several studies have used ML to downscale physical quantities such as precipitation \cite{Vandal_2017:DeepDS}, but without quantifying the prediction uncertainty. Yet, a generally applicable method to downscale arbitrary variables is lacking.

Another difficulty is to remove the bias in the low resolution data. This is an instance of domain adaptation, a topic popularly studied in computer vision. Recent work has used generative models such as GANs and diffusion models to bridge the gap between two domains~\cite{bischoff2022:unpaired,bortoli2021diffusion,choi2021ilvr,meng2021sdedit,pan2021learning,park2020contrastive,sasaki2021:UNIT-DDPM,su2023:dual_diffusion,wu2022unifying,zhao2022egsde}.  A popular domain alignment method that was used in \cite{climalign:2021} for downscaling weather data is AlignFlow \cite{grover2019alignflow}. This approach learns normalizing flows for source and target data of the same dimension, and uses their common latent space to move across domains. The advantage of those methods is that they do not require training data from two domains in correspondence. Many of those approaches are related to optimal transport (OT), a rigorous mathematical framework for learning maps between two domains without paired data~\cite{villani2009optimal}. Recent computational advances in OT for discrete (i.e., empirical) measures~\cite{cuturi2013sinkhorn,peyre2019computational} have resulted in a wide set of methods for domain adaptation~\cite{courty2017joint,flamary2016optimal}. Despite their empirical success with careful choices of regularization, their use alone for high-dimensional images has remained limited~\cite{papadakis2015optimal}.

Our work uses diffusion models to perform upsampling after a debiasing step implemented with OT. We avoid common issues from GANs~\cite{tian2022generative} and flow-based methods~\cite{lugmayr2020srflow}, which include over-smoothing, mode collapse and large model footprints~\cite{dhariwal2021diffusion,Li2022:SRDiff}. Also, due to the debiasing map, which matches the low-frequency content in distribution (see Fig.~\ref{fig:framework_diagram}(a)), we do not need to explicitly impose that the low-frequency power spectra of the two datasets match like some competing methods do~\cite{bischoff2022:unpaired}. Compared to formulations that perform upsampling and debiasing simultaneously~\cite{bischoff2022:unpaired,Vandal_2017:DeepDS}, our framework performs these two tasks separately, by only training (and independently validating) a single probabilistic diffusion model for the high-resolution data once. This allows us to quickly assess different modeling choices, such as the linear downsampling map, by combining the diffusion model with different debiasing maps. Lastly, in comparison to other two-stage approaches~\cite{bischoff2022:unpaired,climalign:2021}, debiasing is conducted at low-resolutions, which is less expensive as it is performed on a much smaller space, and more efficient as it is not hampered from spurious biases introduced by interpolation techniques. 


\else 
\section{Related work}
\label{sRelated}




\textbf{\fs{A lot of methods are enumerated but the texts do not say how our work is contextualized: how about none of those aims to match distributions? Just to find a high-level aspect to contrast our method to} Classic downscaling methods:} 
We broadly categorize methods for super-resolution into two groups: generation of paired data, and 
denoising techniques that don't require paired data. The generation of paired data has been a practical approach, particularly in the engineering community. These methods carefully manipulate high and low resolution models, either by nudging or by enforcing boundary conditions, to produce paired data without introducing spectral biases
~\cite{charalampopoulos2023statistics,dixon2016evaluating}. On the other hand, optimization methods solve an inverse problem to directly estimate the high-resolution data by assuming strong prior knowledge about the recovered signals. For instance, 
convex programming methods developed in the applied mathematics community~\cite{Candes:2013,Candes_Fernandez_Granda_2014:Super_resolution} leverage analytical properties 
such as sparsity in the Fourier domain~\cite{Donoho:super_resolution1992} or translation invariance~\cite{Hua_Sarkar:1991}. Even though such methods have been widely applied for signal processing, their applicability to 
image data has been limited. Finally, the climate/weather community has proposed several custom estimators~\cite{Wilby_1998:downscaling}. Among them, perhaps the most predominant have been bias-correction spatial disaggregation (BCSD), which combines traditional spline interpolation with a quantile matching bias correction \cite{Maraun2013:quantile_matching}, and linear models~\cite{Hessami_2008:linear_models_statistical_dowscaling}.
Recently, several studies have used ML to downscale physical quantities such as precipitation \cite{Vandal_2017:DeepDS}, but without quantifying the prediction uncertainty. See \cite{vandal2017intercomparison} for an extensive overview in weather/climate.

\textbf{\fs{Again, need to contextualize our work within this body of work.}Unpaired image-to-image translation and domain adaptation:} 
In machine learning, several generative models have been developed to sample images that preserves features of a reference image. 
For example,~\cite{park2020contrastive} uses a GAN-based contrastive learning approach,~\cite{pan2021learning} proposes a framework reminiscent of cycle-GANs~\cite{Zhu2017:CycleGAN} that is regularized by matching physically relevant statistics, while~\cite{choi2021ilvr, zhao2022egsde} use diffusion models, and \cite{meng2021sdedit,su2023:dual_diffusion} use a Schr\"odinger brigde instantiated by two diffusion models~\cite{bortoli2021diffusion}. The latter formalism has been applied to statistical downscaling in \cite{bischoff2022:unpaired}. Other approaches that require learning two different diffusion models for the source and target images include 
\cite{wu2022unifying}, and \cite{sasaki2021:UNIT-DDPM}, which alternates between the score functions of two models during the sampling process. A popular domain alignment method that was used in \cite{climalign:2021} for downscaling weather data is AlignFlow \cite{grover2019alignflow}. This approach learns normalizing flows for source and target data of the same dimension, and uses their common latent space to move across domains. 
Lastly, a rigorous mathematical framework for learning maps between two domains without paired data is optimal transport (OT)~\cite{villani2009optimal}. Recent computational advances in OT for discrete (i.e., empirical) measures~\cite{peyre2019computational, cuturi2013sinkhorn} have resulted in a wide set of methods for domain adaptation~\cite{courty2017joint, flamary2016optimal}. Despite their empirical success with careful choices of regularization, 
their use alone for high-dimensional images has remained limited~\cite{papadakis2015optimal}. 

\textbf{Differences with our contribution:} \fs{I think this comparison is clear -- so perhaps using the texts here as backbone, and weave thru other references in the preceding two paragraphs around this text.}By using diffusion models, we avoid common issues from GANs~\cite{tian2022generative} and flow-based methods, which include over-smoothing, mode collapse and large model footprints~\cite{Li2022:SRDiff}. Also, due to the debiasing map, which matches the low-frequency content in distribution (see Fig.~\ref{fig:framework_diagram} (a)), we do not need to explicitly impose that the low-frequency power spectra of the two datasets match as competing methods~\cite{bischoff2022:unpaired}. Compared to formulations that perform upsampling and debias simultaneously~\cite{bischoff2022:unpaired,Vandal_2017:DeepDS}, our framework performs these two tasks separately, by only training (and independently validating) a single probabilistic diffusion model for the high-resolution data once. This allows us to quickly assess different modeling choices, such as the linear down-sampling map, by combining the diffusion model with different debiasing maps. Also, compared to other two-stage approaches~\cite{climalign:2021,bischoff2022:unpaired}, the debiasing is performed at low-resolution, which is less expensive, as it is performed on a much smaller space, and more efficient, as it is not hampered by the spurious biases introduced by interpolation techniques.


\fi
\section{Methodology}
\label{sec:method}

\label{sec:main_idea}

\paragraph{Setup} We consider two spaces: the high-fidelity, high-resolution space $\mathcal{X} = \mathbb{R}^d$ and the low-fidelity, low-resolution space $\mathcal{Y} = \mathbb{R}^{d'}$, where we suppose that $d > d'$. We model the elements $X \in \mathcal{X}$ and $Y\in \mathcal{Y}$ 
as random variables with marginal distributions, $\mu_X$ and $\mu_Y$, respectively. In addition, we suppose there is a 
statistical model relating the $X$ and $Y$ variables 
via $C\colon \mathcal{X} \rightarrow \mathcal{Y}$, an unknown and possibly nonlinear, downsampling map. See Fig.~\ref{fig:framework_diagram}(b) for a diagram. 

Given an observed realization $\bar{y} \in \mathcal{Y}$, which we refer to as a \emph{snapshot}, we formulate downscaling as the problem of sampling from the conditional probability distribution $p(x | E_{\bar{y}})$ for the event $E_{\bar{y}}: = \{ x \in \mathcal{X} \, | \, C(x) = \bar{y}\}$, 
which we denote by  $p(x | C(x) = \bar{y})$.
Our objective is to sample this distribution given only access to marginal samples of $X$ and $Y$.

\paragraph{Main idea} In general, downscaling is an ill-posed problem given that the joint distribution of $X$ and $Y$ is not prescribed by a known statistical model. Therefore, we seek an approximation to $C$ so the statistical properties of $X$ are preserved given samples of $\mu_Y$. In particular, such a map should satisfy $C_{\sharp} \mu_X = \mu_Y$, where $C_{\sharp} \mu_X$ denotes the push-forward measure of $\mu_X$ through $C$.

In this work, we impose a structured ansatz to approximate $C$. Specifically, we \emph{factorize} the map $C$ as the composition of a known and linear \textit{downsampling map} $C'$, and an invertible \textit{debiasing map} $T$:
\begin{equation}
\label{eq:factorization}
    C = T^{-1} \hspace{-2pt}\circ C', \quad \text{such that} \quad (T^{-1} \hspace{-2pt}\circ C')_{\sharp} \mu_X = \mu_Y, 
\end{equation}
or alternatively, $C'_{\sharp} \mu_X = T_{\sharp} \mu_Y$.
This factorization decouples and explicitly addresses two entangled goals in downscaling:  debiasing and upsampling. We discuss the advantage of such factorization, after sketching how $C'$ and $T$ are implemented.

The range of the downsampling map $C' \colon \mathcal{X} \rightarrow \mathcal{Y}'$ defines an \emph{intermediate} space $\mathcal{Y}' = \mathbb{R}^{d'}$ of high-fidelity low-resolution samples with measure $\mu_{Y'}$. Moreover, the joint space $\mathcal{X} \times \mathcal{Y}'$ is built by projecting samples of $X$ into $\mathcal{Y}'$, i.e., $(x, y') = (x, C'x) \in \mathcal{X}\times \mathcal{Y}'$; see Fig.~\ref{fig:framework_diagram}(b). Using these spaces, we decompose the domain adaptation problem into the following three sub-problems:
\begin{enumerate}
    \item \textit{High-resolution prior}: Estimate the marginal density $p(x)$;
    \item \textit{Conditional modeling}: For the joint variables ${X} \times {Y}'$, approximate $p(x | C'x = y')$;
    \item \textit{Debiasing}: Compute a transport map such that $T_{\sharp} \mu_Y = C'_{\sharp} \mu_X$. 
\end{enumerate}

For the first sub-problem, we train an \textit{unconditional} model to approximate $\mu_X$, or $p(x)$, as explained in \S\ref{sec:diffusion}. For the second sub-problem, we leverage the prior model and $y' \in \mathcal{Y}'$ to build a model for \textit{a posteriori} conditional sampling of $p(x | C'x = y')$, which allows us to upsample snapshots from $ \mathcal{Y}'$ to $ \mathcal{X}$, as explained in \S\ref{sec:conditional_diffusion}.
For the third sub-problem, we use domain adaptation to shift the resulting model from the source domain $ \mathcal{X}\times  \mathcal{Y}'$ to the target domain $\mathcal{X} \times \mathcal{Y}$, for which there is no labeled data. For such a task, we build a transport map $T:  \mathcal{Y} \rightarrow  \mathcal{Y}'$ satisfying the condition that $T_{\sharp} \mu_Y = \mu_{Y'} = C'_{\sharp} \mu_X$. This map is found by solving an optimal transport problem, which we explain in \S\ref{sec:OTmaps}. 

Lastly, we merge the solutions to the sub-problems to arrive at our core downscaling methodology, which is summarized in Alg. \ref{alg:inference}. In particular, given a low-fidelity and low-resolution sample $\overline{y}$, we use the optimal transport map $T$ to project the sample to the high-fidelity space $\overline{y}'\hspace{-1pt} = \hspace{-1pt} T(\overline{y})$ and use the conditional model to sample $p(x|C'x \hspace{-1pt}= \hspace{-1pt} \overline{y}')$. The resulting samples are contained in the high-fidelity and high-resolution space.

The factorization in \eqref{eq:factorization} has several advantages. We do not require a cycle-consistency type of loss~\cite{grover2019alignflow,Zhu2017:CycleGAN}: the consistency condition is automatically enforced by \eqref{eq:factorization} and the conditional sampling. By using a linear downsampling map $C'$, it is trivial to create the intermediate space $\mathcal{Y}'$, while rendering the conditional sampling tractable: conditional sampling with a nonlinear map is often more expensive and it requires more involved tuning~\cite{chung2022diffusion,chung2022improving}.
The factorization also allows us to compute the debiasing map in a considerably lower dimensional space, which conveniently requires less data to cover the full distribution, and fewer iterations to find the optimal map~\cite{cuturi2013sinkhorn}. 


\begin{algorithm}[t]
\caption{: \textbf{Downscaling from $\bar{y} \in \mathcal{Y}$ to $\bar{x}(\bar{y}) \in \mathcal{X}$.}}
\label{alg:inference}
\begin{algorithmic}
\State \textbf{Input: low-fidelity, low-resolution sample: $y$}
\State 1. Compute the debiased term  $\bar{y}' = T_{\gamma}(\bar{y})$ using the barycentric approximation in \eqref{eq:barycentric_OT}.
\State 2. Modify the scoring function in \eqref{eq:denoiser_to_score} using the conditional denoiser in \eqref{eq:conditioned_denoiser}.
\State 3. Solve the reverse SDE in \eqref{eq:bwd_sde}, and obtain $\bar{x}$.
\State \textbf{Output: high-fidelity, high-resolution sample :} $\bar{x}$
\end{algorithmic}
\end{algorithm}

\subsection{High-resolution prior}
\label{sec:diffusion}
To approximate the prior of the high-resolution snapshots we use a probabilistic diffusion model, which is known to avoid several drawbacks of other generative models used for super-resolution~\cite{Li2022:SRDiff}, while providing greater flexibility for \emph{a posteriori} conditioning \cite{chung2022diffusion,pmlrfinzi23a,kawar2021snips,kawar2022denoising}.  

Intuitively, diffusion-based generative models involves iteratively transforming samples from an initial noise distribution $p_T$ into ones from the target data distribution $p_0 = p_{\text{data}}$. Noise is removed sequentially such that samples follow a family of marginal distributions $p_t(x_t; \sigma_t)$ for decreasing diffusion times $t$ and noise levels $\sigma_t$. Conveniently, such distributions are given by a forward noising process that is described by the stochastic differential equation (SDE)~\cite{karras2022elucidating,song2020score}
\begin{equation}
\label{eq:fwd_sde}
  dx_t  = f(x_t, t) dt +g(x_t, t) dW_t, 
\end{equation}
with drift $f$, diffusion coefficient $g$, and the standard Wiener process $W_t$. Following~\cite{karras2022elucidating}, we set
\begin{equation}
    f(x_t, t) = f(t) x_t := \frac{\dot{s}_t}{s_t}x_t,  \qquad \text{and} \qquad
    g(x_t, t) = g(t) := s_t\sqrt{2\dot{\sigma}_t\sigma_t}.
\end{equation}
Solving the SDE in~\eqref{eq:fwd_sde} forward in time with an initial condition $x_0$ leads to the Gaussian perturbation kernel $p(x_t|x_0) = \mathcal{N}(x_t; s_tx_0, s_t^2\sigma_t^2\textbf{I})$. Integrating the kernel over the data distribution $p_0(x_0) = p_{\text{data}}$, we obtain the marginal distribution $p_t(x_t)$ at any $t$. As such, one may prescribe the profiles of $s_t$ and $\sigma_t$ so that $p_0 = p_{\text{data}}$ (with $s_0 = 1, \sigma_0 = 0$),
and more importantly
\begin{equation}
\label{eq:dist_T}
    p_T(x_T) \approx \mathcal{N}(x_T; 0, s_T^2\sigma_T^2\mathbf{I}),
\end{equation}
i.e., the distribution at the terminal time $T$ becomes indistinguishable from an isotropic, zero-mean Gaussian. To sample from $p_{\text{data}}$, we utilize the fact that the  
reverse-time SDE 
\begin{equation}
\label{eq:bwd_sde}
    dx_t = \big[f(t) x_t - g(t)^2\nabla_{x_t} \log p_t(x_t)\big]dt +g(t) dW_t,
\end{equation}
has the same marginals as \eqref{eq:fwd_sde}. Thus, by solving \eqref{eq:bwd_sde} backwards using \eqref{eq:dist_T} as the final condition at time $T$, we obtain samples from $p_{\text{data}}$ at $t=0$.


Therefore, the problem is reduced to estimating the \emph{score function} $\nabla_{x_t}\log{p_t(x_t)}$ resulting from $p_\text{data}$ and the prescribed diffusion schedule $(s_t, \sigma_t)$. We adopt the denoising formulation in~\cite{karras2022elucidating} and learn a neural network $D_\theta(x_0 + \varepsilon_t, \sigma_t)$, where $\theta$ denotes the network parameters. The learning seeks to minimize the $L_2$-error in predicting the true sample $x_0$ given a noise level $\sigma_t$ and the sample noised with $\varepsilon_t = \sigma_t\varepsilon$ where $\varepsilon$ is drawn from a standard Gaussian. The score can then be readily obtained from the denoiser $D_\theta$ via the asymptotic relation (i.e., Tweedie's formula~\cite{efron2011tweedie})
\begin{equation}
\label{eq:denoiser_to_score}
    \nabla_{x_t}\log{p_t(x_t)} \approx \frac{D_\theta(\hat{x}_t, \sigma_t) - \hat{x}_t}{s_t\sigma_t^2}, \qquad \hat{x}_t = x_t / s_t.
\end{equation}

\subsection{\emph{A posteriori} conditioning via post-processed denoiser}
\label{sec:conditional_diffusion}
We seek to super-resolve a low-resolution snapshot $\bar{y}' \in \mathcal{Y}'$ to a high-resolution one by leveraging the high-resolution prior modeled by the diffusion model introduced above.
Abstractly, our goal is to sample from $p(x_0|E'_{\bar{y}'})$, where $E'_{\bar{y}'}= \{x_0: C' x_0 \hspace{-2pt}=\hspace{-1pt}\bar{y}'\}$. Following \cite{pmlrfinzi23a}, this may be approximated by modifying the learned denoiser $D_\theta$ at \emph{inference time} (see Appendix~\ref{app:conditioning} for more details):
\begin{equation}
\label{eq:conditioned_denoiser}
    \tilde{D}_{\theta}(\hat{x}_t, \sigma_t) = (C')^\dag \bar{y}' + (I - VV^T)\left[D_\theta(\hat{x}_t, \sigma_t) - \alpha\nabla_{\hat{x}_t} \|C' D_\theta(\hat{x}_t, \sigma_t)- \bar{y}'\|^2\right],
\end{equation}
where $(C')^\dag = V\Sigma^{-1}U^T$ is the pseudo-inverse of $C'$ based on its singular value decomposition (SVD) $C'=U\Sigma V^T$, and $\alpha$ is a hyperparameter that is empirically tuned. The $\tilde{D}_{\theta}$ defined in \eqref{eq:conditioned_denoiser} directly replaces $D_\theta$ in~\eqref{eq:denoiser_to_score} to construct a conditional score function $\nabla_{x_t}\log{p_t(x_t|E'_{\bar{y}'})}$ that facilitates the sampling of $p(x_0|E'_{\bar{y}'})$ using the reverse-time SDE in~\eqref{eq:bwd_sde}.

\subsection{Debiasing via optimal transport} \label{sec:OTmaps}

In order to upsample a biased low-resolution data $\overline{y} \in \mathcal{Y}$, we first seek to find a mapping $T$ such that $\overline{y}' = T(\overline{y}) \in \mathcal{Y}'$ is a representative sample from the distribution of unbiased low-resolution data. Among the infinitely many maps that satisfy this condition, the framework of optimal transport (OT) selects a map by minimizing an integrated transportation distance based on the cost function $c \colon \mathcal{Y} \times \mathcal{Y}' \rightarrow \R^{+}$. The function $c(y,y')$ defines the cost of moving one unit of probability mass from $y'$ to $y$. By treating $Y,Y'$ as random variables on $\mathcal{Y}, \mathcal{Y'}$ 
with measures $\mu_{Y},\mu_{Y'}$, respectively, the OT map is given by the solution to the Monge problem
\begin{equation} \label{eq:MongeProblem}
    \min_{T} \left\{ \int c(y,T(y))d\mu_{Y}(y): T_\sharp \mu_Y = \mu_{Y'} \right\}.
\end{equation}

In practice, directly solving the Monge problem is hard and may not even admit a solution~\cite{villani2009optimal}. One common relaxation of \eqref{eq:MongeProblem} is to seek a joint distribution, known as a coupling or transport plan, which relates the underlying random variables~\cite{villani2009optimal}. 
A valid plan is a probability measure $\gamma$ on $ \mathcal{Y} \times \mathcal{Y}'$ with marginals $\mu_{Y}$ and $\mu_{Y'}$.
To efficiently estimate the plan when the $c$ is the quadratic cost (i.e., $c(y, y') = \frac{1}{2}\|y - y' \|^2$), we solve the entropy regularized problem 
\begin{equation} \label{eq:EntropyReg_Wasserstein}
    \inf_{\gamma \in \Pi(\mu_Y,\mu_{Y'})} \int \frac{1}{2}\|y-y'\| ^2 d\gamma(y,y') + \epsilon D_{\textrm{KL}}(\gamma||\mu_{Y'} \otimes \mu_Y),
\end{equation}
where $D_{\text{KL}}$ denotes the KL divergence, and $\epsilon > 0$ is a small regularization parameter, using the Sinkhorn's algorithm \cite{cuturi2013sinkhorn}, which leverages the structure of the optimal plan to solve~\eqref{eq:EntropyReg_Wasserstein} with small runtime complexity~\cite{altschuler2017near}. The solution to \eqref{eq:EntropyReg_Wasserstein} is the transport plan $\gamma_\epsilon \in \Pi(\mu,\nu)$ given by
\begin{equation} \label{eq:EntropyReg_Plan}
  \gamma_\epsilon(y,y') = \exp\left((f_\epsilon(y) + g_\epsilon(y') - \frac{1}{2}\|y - y'\|^2)/\epsilon\right)d\mu_Y(y)d\mu_{Y'}(y'),  
\end{equation}
in terms of potential functions $f_\epsilon, g_\epsilon$ that are chosen to satisfy the marginal constraints. After finding these potentials, we can approximate the transport map using the barycentric projection $T_\gamma(y) = \mathbb{E}_{\gamma}[Y'|Y = y],$ 
for a plan $\gamma \in \Pi(\mu_Y,\mu_{Y'})$ \cite{Agueh2011:Barycenter_map}. 
For the plan in~\eqref{eq:EntropyReg_Plan}, the map is given by
\begin{equation}\label{eq:barycentric_OT}
    T_{\gamma_\epsilon}(y) = \frac{\int y' e^{(g_\epsilon(y') -\frac{1}{2}\|y - y'\|^2)/\epsilon} d\mu_Y(y')}{\int e^{(g_\epsilon(y') -\frac{1}{2}\|y - y'\|^2)/\epsilon} d\mu_Y(y')}.
\end{equation}
In this work, we estimate the potential functions $f_\epsilon,g_\epsilon$ from samples, i.e., empirical approximations of the measures $\mu_Y,\mu_{Y'}$. Plugging in the estimated potentials in~\eqref{eq:barycentric_OT} defines an approximate transport map to push forward samples of 
$\mu_{Y}$ to $\mu_{Y'}$. More details on the estimation of the OT map are provided in Appendix~\ref{app:debiasing}.


A core advantage of this methodology is that it provides us with the flexibility of changing the cost function $c$ in~\eqref{eq:MongeProblem}, and embed it with structural biases that one wishes to preserve in the push-forward distribution. Such direction is left for future work. 



\section{Numerical experiments}
\label{sec:numerical_experiments}

\subsection{Data and setup}
\label{sec:data}

We showcase the efficacy and performance of the proposed approach on one- and two-dimensional fluid flow problems that are representative of the core difficulties present in numerical simulations of weather and climate. We consider the one-dimensional Kuramoto-Sivashinski (KS) equation and the two-dimensional Navier-Stokes (NS) equation under Kolmogorov forcing (details in Appendix~\ref{app:datasets}) in periodic domains. The low-fidelity (LF), low-resolution (LR) data ($\mathcal{Y}$ in Fig.~\ref{fig:framework_diagram}(b)) is generated using a finite volume discretization in space~\cite{leveque_2002} and a fractional discretization in time, while the high-fidelity (HF), high-resolution (HR) data ($\mathcal{X}$ in Fig.~\ref{fig:framework_diagram}(b)) is simulated using a spectral discretization in space with an implicit-explicit scheme in time. 
Both schemes are implemented with \texttt{jax-cfd} and its finite-volume and spectral toolboxes~\cite{Dresdner:2020ml_spectral,kochkov_machine_2021} respectively.
After generating the HF data in HR, we run the LF solver using a spatial discretization that is $8\times$ coarser (in each dimension) with permissible time steps. For NS, we additionally create a $16\times$ coarser LFLR dataset by further downsampling by a factor of two the $8\times$ LFLR data. See Appendix~\ref{app:datasets} for further details. 

For both systems, the datasets consist of long trajectories generated with random initial conditions\footnote{The presence of global attractors in both systems renders the exact initial conditions unimportant. It also guarantees sufficient coverage of the target distributions sampling from long trajectories.}, which are sufficiently downsampled in time to ensure that consecutive samples are decorrelated. We stress once more that even when the grids and time stamps of both methods are aligned, there is \emph{no pointwise correspondence} between elements of $\mathcal{X}$ and $\mathcal{Y}$. This arises from the different modeling biases inherent to the LF and HF solvers, which inevitably disrupt any short-term correspondence over the long time horizon in a strongly nonlinear dynamical setting. 

Finally, we create the intermediate space $\mathcal{Y}'$ in Fig.~\ref{fig:framework_diagram}(b) by downsampling the HFHR data with a simple selection mask\footnote{It is worth noting that careful consideration should be given to the choice of $C'$ to avoid introducing aliasing, as this can potentially make the downscaling task more challenging.} (i.e., the map $C'$).  
This creates the new HFLR dataset $\mathcal{Y}'$ with the same resolution as $\mathcal{Y}$, 
but with the low-frequency bias structure of $\mathcal{X}$ induced by the push-forward of $C'$. 

\textbf{Baselines and definitions.} 
We define the following ablating variants of our proposed method
\begin{itemize}
    \item Unconditional diffusion sampling (\textit{UncondDfn}).
    \item Diffusion sampling conditioned on LFLR data without OT correction (\textit{Raw cDfn}).
    \item {[\textit{Main}]} Diffusion sampling conditioned on OT-corrected (HFLR) data (\textit{OT+cDfn}).
\end{itemize}
We additionally consider the following baselines to benchmark our method:
\begin{itemize}
    \item Cubic interpolation approximating HR target using local third-order splines  (\textit{Cubic}).
    \item Vision transformer (\textit{ViT})~\cite{dosovitskiy2021an} based deterministic super-resolution model.
     \item Bias correction and statistical disaggregation (\textit{BCSD}), involving upsampling with cubic interpolation, followed by a quantile-matching debiasing step.
    \item CycleGAN, which is adapted from~\cite{Zhu2017:CycleGAN} to enable learning transformations between spaces of different dimensions (\textit{cycGAN}).
    \item ClimAlign (adapted from \cite{climalign:2021}), in which the input is upsampled using cubic interpolation, and the debiasing step is performed using AlignFlow \cite{grover2019alignflow} (\textit{ClimAlign}).
\end{itemize}
The first two baselines require paired data and, therefore, learn the upsampling map $\mathcal{Y}'\rightarrow\mathcal{X}$ (i.e., HFLR to HFHR) and are composed with OT debiasing as factorized baselines. BCSD is a common approach used in the climate literature.
The last two baselines present end-to-end alternatives and are trained directly on unpaired LFLR and HFHR samples. Further information about the implemented baselines can be found in Appendix~\ref{app:baselines}.

\textbf{OT training.}  
To learn the transport map in \eqref{eq:barycentric_OT}, we solve the entropic OT problem in~\eqref{eq:EntropyReg_Wasserstein} with $\epsilon = 0.001$ using a Sinkhorn \cite{cuturi2013sinkhorn} iteration with Anderson acceleration and parallel updates. We use $90,000$ i.i.d.\thinspace samples of $Y \in \mathcal{Y}$ and $Y' \in \mathcal{Y}'$, and perform $5000$ iterations. Implementations are based on the \texttt{ott-jax} library \cite{cuturi2022ott}. 

\textbf{Denoiser training and conditional sampling.} 
The denoiser $D_\theta$ is parametrized with a standard U-Net architecture similar to the one used in~\cite{saharia2022photorealistic}. We additionally incorporate the preconditioning technique proposed in~\cite{karras2022elucidating}.
For $s_t$ and $\sigma_t$ schedules, we employ the variance-preserving (VP) scheme originally introduced in~\cite{song2020score}. Furthermore, we adopt a data augmentation procedure to increase the effective training data size by taking advantage of the translation symmetries in the studied systems. 

Samples are generated by solving the SDE based on the post-processed denoiser $\tilde{D}_\theta$ using the Euler-Maruyama scheme with exponential time steps, i.e., $\{t_i\}$ is set such that $\sigma(t_i) = \sigma_{\text{max}}(\sigma_{\text{min}}/\sigma_{\text{max}})^{i/N}$ for $i = \{0, ..., N\}$. The number of steps used, $N$, vary between systems and downscaling factors. More details regarding denoiser training and sampling are included in Appendix~\ref{app:diffusion}.

\textbf{Metrics.} 
To quantitatively assess the quality of the resulting snapshots we compare a number of physical and statistical properties of the snapshots: (i) the energy spectrum, which measures the energy in each Fourier mode and thereby providing insights into the similarity between the generated and reference samples, (ii) a spatial covariance metric, which characterizes the spatial correlations within the snapshots, (iii) the KL-divergence (KLD) of the kernel density estimation for each point, which serves as a measure for the local structures (iv) the maximum mean discrepancy (MMD), and (v) the empirical Wasserstein-1 metric (Wass1). We present (i) below and leave the rest described in Appendix~\ref{app:metrics} as they are commonly used in the context of probabilistic modeling. 

The energy spectrum is defined\footnote{This definition is applied to each sample and averaged to obtain the metric (same for MELR below).} as 
\begin{equation}
\label{eq:energy_spectrum}
    E(k) = \sum_{|\underline{k}| =  k} | \hat{u}(\underline{k}) | ^2 = \sum_{|\underline{k}| = k} \left | \sum_{i} u(x_i) \exp(-j 2\pi \underline{k} \cdot x_i/L) \right|^2
\end{equation}
where $u$ is a snapshot system state, and $k$ is the magnitude of the wave-number (wave-vector in 2D) $\underline{k}$. 
To assess the overall consistency of the spectrum between the generated and reference samples using a single scalar measure, we consider the mean energy log ratio (MELR):
\begin{equation}
\label{eq:MELR}
    \text{MELR} = \sum_k w_k\left |\log \left (E_{\text{pred}}(k) /E_{\text{ref}}(k) \right )\right |,
\end{equation}
where $w_k$ represents the weight assigned to each $k$. We further define $w_{k}^{\text{unweighted}} = 1/\text{card}(k)$ and $w_{k}^{\text{weighted}} = E_{\text{ref}}(k)/\sum_k E_\text{ref}(k)$. The latter skews more towards high-energy/low-frequency modes.

\subsection{Main results}
\label{sec:main_results}

\begin{figure}[t]
  \centering
  {\setlength\tabcolsep{0.5pt} {\renewcommand{\arraystretch}{0.5}
      \begin{tabular}{M{.32\linewidth}M{.32\linewidth}M{.32\linewidth}}
        \hspace{-3pt}\includegraphics[width=\linewidth]{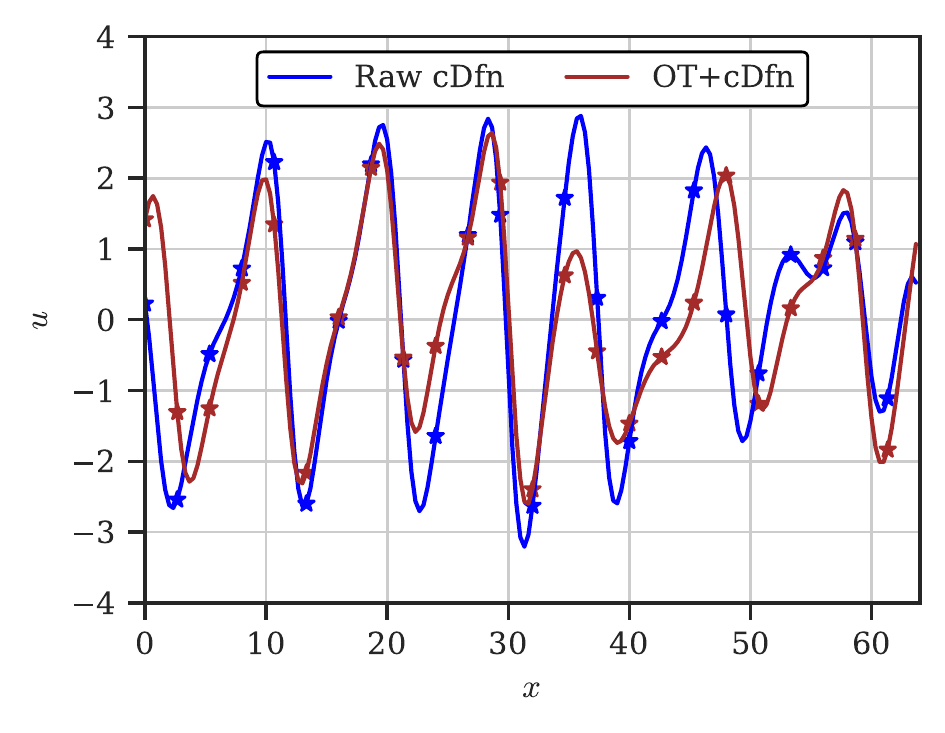} &
        \hspace{-3pt}\includegraphics[width=\linewidth]{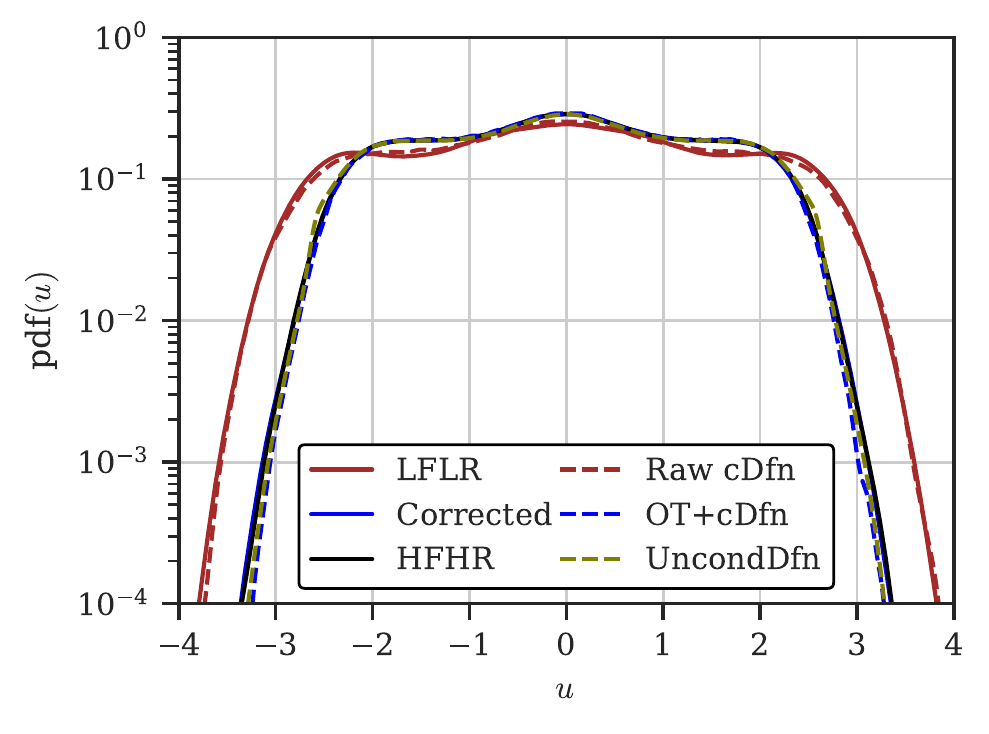} & 
        \hspace{-3pt}\includegraphics[width=\linewidth]{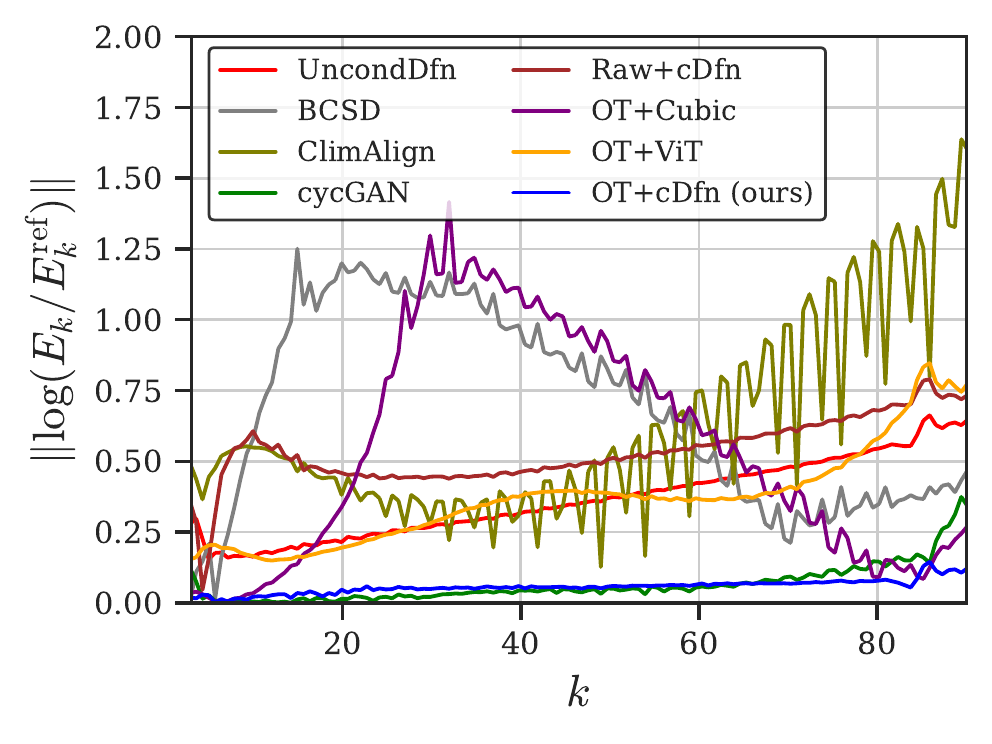} \\
        (a) KS, conditional samples & \quad (b) KS, marginal PDFs & \quad (c) NS $8\times$, log energy ratios \\
      \end{tabular}
  }}
  \caption{(a) KS samples generated with diffusion model conditioned on LR information with and without OT correction applied, (b) empirical probability density function for relevant LR and HR samples in KS and (c) mode-wise log energy ratios with respect to the true samples (\eqref{eq:MELR} without weighted sum) at $8\times$ downscaling for NS.}
  \label{fig:combo_plots}
\end{figure}

\begin{table}[t]
\centering
\caption{Metrics of the LFLR source and OT-corrected samples for KS and NS. The precise metric definitions are provided in Appendix~\ref{app:metrics}.}

{\setlength\tabcolsep{3.5pt} {\setlength{\extrarowheight}{2pt}
\begin{tabular}{l|cc|cc|cc|}
\cline{2-7}
                                          & \multicolumn{2}{c|}{KS 8$\times$} & \multicolumn{2}{c|}{NS 8$\times$} & \multicolumn{2}{c|}{NS 16$\times$} \\ \hline
\multicolumn{1}{|l|}{\textbf{Metric}}      & LFLR   & OT-corrected    & LFLR        & OT-corrected         & LFLR         & OT-corrected         \\ \hline
\multicolumn{1}{|l|}{covRMSE $\downarrow$} & $0.343$  & $\mathbf{0.081}$  & $0.458$  & $\mathbf{0.083}$   & $0.477$   & $\mathbf{0.079}$  \\
\multicolumn{1}{|l|}{MELRu $\downarrow$}         & $0.201$  & $\mathbf{0.020}$  & $1.254$  & $\mathbf{0.013}$   & $0.600$   & $\mathbf{0.016}$  \\
\multicolumn{1}{|l|}{MELRw $\downarrow$}           & $0.144$  & $\mathbf{0.020}$  & $0.196$  & $\mathbf{0.026}$   & $0.200$   & $\mathbf{0.025}$  \\
\multicolumn{1}{|l|}{KLD $\downarrow$}                   & $1.464$  & $\mathbf{0.018}$  & $29.30$  & $\mathbf{0.033}$   & $12.26$   & $\mathbf{0.017}$  \\ \hline
\end{tabular}
}}
\label{table:OT_results}
\end{table}


\begin{figure}[h]
\centering
\includegraphics[width=\linewidth]{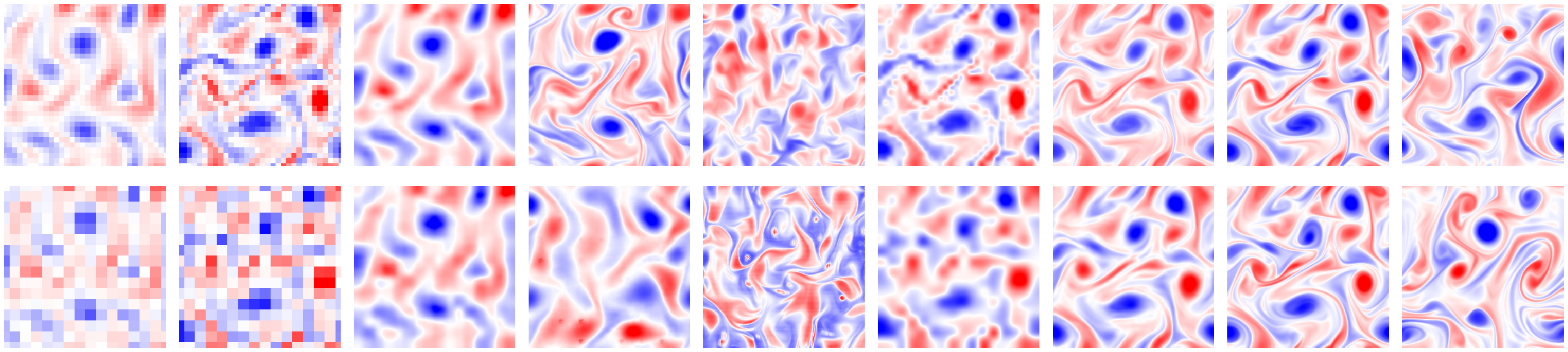}
{\setlength\tabcolsep{0.4pt}
\begin{tabular}{M{.11\linewidth}M{.11\linewidth}M{.11\linewidth}M{.11\linewidth}M{.11\linewidth}M{.11\linewidth}M{.11\linewidth}M{.11\linewidth}M{.11\linewidth}}
     (a) & (b) & (c) & (d) & (e) & (f) & (g)  & (h) & (i)
\end{tabular}
}
\caption{Example showing the vorticity field of samples debiased and super-resolved using different techniques at $8\times$ (top row) and $16\times$ (bottom row) downscaling factors. From left to right: \textbf{(a)} LR snapshots produced by the \textbf{low-fidelity solver} (input $\bar{y}$ of Alg. \ref{alg:inference}), \textbf{(b) OT-corrected} snapshots ($\bar{y}'$ in line $1$ of Alg. \ref{alg:inference}), \textbf{(c) BCSD} applied to LR snapshots, \textbf{(d)} snapshots downscaled with \textbf{cycle-GAN} directly from LR snapshots, \textbf{(e) ClimAlign} applied to LR snapshots, \textbf{(f) cubic interpolation} of the OT-corrected snapshots, \textbf{(g)} deterministic upsample of the OT-corrected snapshots with \textbf{ViT}, \textbf{(h) diffusion sample conditioned on the OT-corrected snapshots} (output $\bar{x}$ in Alg.~\ref{alg:inference}, ours), and \textbf{(i)} two \textbf{true HR samples} in the training data with the closest Euclidean distance to the OT-corrected generated sample. The $16\times$ source is the same as the $8\times$ source but further downsampled by a factor of two. OT maps are computed independently between resolutions.}
\label{fig:sample_comparisons}
\end{figure}

\textbf{Effective debiasing via optimal transport. } 
Table \ref{table:OT_results} shows that the OT map effectively corrects the statistical biases in the LF snapshots for all three experiments considered. Significant improvements are observed across all metrics, demonstrating that the OT map approximately achieves $C'_{\sharp} \mu_X \approx T_{\sharp} \mu_Y$ as elaborated in \S\ref{sec:main_idea} (extra comparisons are included in Appendix~\ref{app:debiasing}). 

Indeed, the OT correction proves crucial for the success of our subsequent conditional sampling procedure: the unconditional diffusion samples may not have the correct energy spectrum (see \textit{UncondDfn} in Fig.~\ref{fig:combo_plots}(c), i.e. suffering from \emph{color shifts} - a known problem for score-based diffusion models~\cite{song2020improved, choi2022perception}. The conditioning on OT corrected data serves as a sparse anchor which draws the diffusion trajectories to the correct statistics at sampling time.
In fact, when conditioned on uncorrected data, the bias effectively pollutes the statistics of the samples (\textit{Raw cDfn} in Table~\ref{table:ns_metrics}). Fig.~\ref{fig:combo_plots}(b) shows that the same pollution is present for the KS case, despite the unconditional sampler being unbiased.

In Appendix~\ref{app:ablation}, we present additional ablation studies that demonstrate the importance of OT correction in the factorized benchmarks. 

\textbf{Comparison vs. factorized alternatives. }
Fig.~\ref{fig:sample_comparisons} displays NS samples generated by all benchmarked methods. Qualitatively, our method is able to provide highly realistic small-scale features. In comparison, we observe that \textit{Cubic} expectedly yields the lowest quality results; the deterministic \textit{ViT} produces samples with color shift and excessive smoothing, especially at $16\times$ downscaling factor. 

Quantitatively, our method outperforms all competitors in terms of MELR and KLD metrics in the NS tasks, while demonstrating consistently good performance in both $8\times$ and $16\times$ downscaling, despite the lack of recognizable features in the uncorrected LR data (Fig.~\ref{fig:sample_comparisons}(a) bottom) in the latter case. Other baselines, on the other hand, experience a significant performance drop. This showcases the value of having an unconditional prior to rely on when the conditioning provides limited information. 

\textbf{Comparison vs. end-to-end downscaling. }
Although the \textit{cycGAN} baseline is capable of generating high-quality samples at $8\times$ downscaling (albeit with some smoothing) reflecting competitive metrics, we encountered persistent stability issues during training, particularly in the $16\times$ downscaling case. 



\textbf{Diffusion samples exhibit ample variability. } 
Due to the probabilistic nature of our approach, we can observe from Table \ref{table:ns_metrics} that the OT-conditioned diffusion model provides some variability in the downscaling task,
which increases when the downscaling factor increases. This variability provides a measure of uncertainty quantification in the generated snapshots as a result of the consistent formulation of our approach on probability spaces.

\begin{table}[t]
\centering
\caption{Evaluation of downscaling methods for NS. The best metric values are highlighted \textbf{in bold}. Precise metric definitions (except MELR, given by \eqref{eq:MELR}) are included in Appendix~\ref{app:metrics}.}
{
{
\setlength{\extrarowheight}{2pt}
\begin{tabular}{lccccccc}
\hline
Model          & Var  & covRMSE↓ & MELRu↓ & MELRw↓ & KLD↓  & Wass1↓ & MMD↓ \\ \hline
\multicolumn{8}{l}{\textbf{8$\times$ downscale}}                                       \\ \hline
BCSD           & 0    & 0.31     & 0.67   & 0.25   & 2.19  & \textbf{0.23}   & 0.10 \\ \hline
cycGAN         & 0    & 0.15     & 0.08   & 0.05   & 1.62  & 0.32   & 0.08 \\ \hline
ClimAlign      & 0    & 2.19     & 0.64   & 0.45   & 64.37 & 2.77   & 0.53 \\ \hline
Raw+cDfn       & 0.27 & 0.46     & 0.79   & 0.37   & 73.16 & 1.04   & 0.42 \\ \hline
OT+Cubic       & 0    & \textbf{0.12}     & 0.52   & 0.06   & 1.46  & 0.42   & 0.10 \\ \hline
OT+ViT         & 0    & 0.43     & 0.38   & 0.18   & 1.72  & 1.11   & 0.31 \\ \hline
(ours) OT+cDfn & 0.36 & \textbf{0.12}     & \textbf{0.06}   & \textbf{0.02}   & \textbf{1.40}  & 0.26   & \textbf{0.07} \\ \hline
\multicolumn{8}{l}{\textbf{16$\times$ downscale}}                                      \\ \hline
BCSD           & 0    & 0.34     & 0.67   & 0.25   & 2.17  & \textbf{0.21}   & 0.11 \\ \hline
cycGAN         & 0    & 0.32     & 1.14   & 0.28   & 2.05  & 0.48   & 0.13 \\ \hline
ClimAlign      & 0    & 2.53     & 0.81   & 0.50   & 77.51 & 3.15   & 0.55 \\ \hline
Raw+cDfn       & 1.07 & 0.46     & 0.54   & 0.30   & 93.87 & 0.99   & 0.39 \\ \hline
OT+Cubic       & 0    & 0.25     & 0.55   & 0.13   & 7.30  & 0.85   & 0.20 \\ \hline
OT+ViT         & 0    & 0.14     & 1.38   & 0.09   & 1.67  & 0.32   & \textbf{0.07} \\ \hline
(ours) OT+cDfn & 1.56 & \textbf{0.12}     & \textbf{0.05}   & \textbf{0.02}   & \textbf{0.83}  & 0.29   & \textbf{0.07} \\ \hline
\end{tabular}
}}
\label{table:ns_metrics}
\end{table}

\section{Conclusion}

We introduced a two-stage probabilistic framework for the statistical downscaling problem. The framework performs a debiasing step to correct the low-frequency statistics, followed by an upsampling step using a conditional diffusion model. We demonstrate that when applied to idealized physical fluids, our method provides high-resolution samples whose statistics are physically correct, even when there is a mismatch in the low-frequency energy spectra between the 
low- and high-resolution data distributions. We have shown that our method is competitive and outperforms several commonly used alternative methods. 

Future work will consider fine-tuning transport maps by adapting the map to the goal of conditional sampling, and introducing physically-motivated cost functions in the debiasing map. Moreover, we will address current limitations of the methodology, such as the high-computational complexity of learning OT-maps that scales quadratically with the size of the training set, and investigate the model's robustness to added noise in the collected samples as is found in weather and climate datasets. We will also further develop this methodology to cover other downscaling setups such as perfect prognosis \cite{maraun_widmann_2018} and spatio-temporal downscaling.


\section*{Broader impact}

Statistical downscaling is important to weather and climate modeling. In this work, we propose a new method for improving the accuracy of high-resolution forecasts (on which risk assessment would be made) from low resolution climate modeling.  Weather and climate research and other scientific communities in computational fluid dynamics will benefit from this work for its potential to reduce computational costs. We do not believe this research  will disadvantage anyone. 

\section*{Acknowledgments}

The authors would like to sincerely thank Toby Bischoff,  Katherine Deck, Nikola Kovachki, Andrew Stuart and Hongkai Zheng for many insightful and inspiring discussions that were vital to this work. RB gratefully acknowledges support from the Air Force Office of Scientific Research MURI on “Machine Learning and Physics-Based Modeling and Simulation” (award FA9550-20-1-0358), and a Department of Defense (DoD) Vannevar Bush Faculty Fellowship (award N00014-22-1-2790).


\bibliography{references}
\bibliographystyle{abbrvnat}
\newpage
\appendix

    


\section{Constrained sampling via post-processed denoiser}
\label{app:conditioning}
In this section, we provide more details on the apparatus necessary to perform \emph{a posteriori} conditional sampling in the presence of a linear constraint.

\eqref{eq:denoiser_to_score} suggests that the SDE drift corresponding to the score may be broken down into 3 steps: 
\begin{enumerate}
    \item The denoiser output $D_\theta(\hat{x}_t, \sigma_t)$ provides a \textit{target sample} that is estimated to be a member of the true data distribution;
    \item The current state $\hat{x}_t$ is \textit{nudged} towards this target\footnote{In the same way that $\dot{a} = -\beta(a_0 - a)$ results in $a\rightarrow a_0$ as $t\rightarrow-\infty$ for any $\beta>0$};
    \item Appropriate rescaling is applied based on the scale and noise levels ($s_t$ and $\sigma_t$) prescribed for the current diffusion time $t$.
\end{enumerate}

For conditional sampling, consider imposing a linear constraint $Cx_0 = y$, where $C\in\mathbb{R}^{d\times d_c}$ and $y\in\mathbb{R}^{d_c}$. Decomposing $C$ in terms of its singular value decomposition (SVD), $C = U\Sigma V^T$, leads to
\begin{equation}
\label{eq:projected_constraint}
    V^Tx_0 = \Sigma^{-1}U^Ty := \tilde{y}.
\end{equation}
Note that this constraint can be easily embedded in the target by replacing the corresponding components of $D_\theta(\hat{x}_t, \sigma_t)$ in this subspace with the constrained value, yielding a post-processed target
\begin{equation}
\label{eq:pp_target_1}
    D_{\theta, \text{cons}}(\hat{x}_t, \sigma_t) = V\tilde{y} + (I - VV^T)D_\theta(\hat{x}_t, \sigma_t).
\end{equation}
This modification alone \textit{guarantees} that the sample $x_0$ produced by the SDE satisfies the required constraint (up to solver errors), while components in the orthogonal complement of the constraint are guided by the denoiser just as in the unconstrained case.

However, in practice this modification creates a "discontinuity" between the constrained and unconstrained components, leading to erroneous correlations between them in the generated samples. As a remedy, we introduce an additional correction in the unconstrained subspace:
\begin{equation}
\label{eq:pp_target_2}
    \tilde{D}_{\theta, \text{cons}}(\hat{x}_t, \sigma_t) = D_{\theta, \text{cons}}(\hat{x}_t, \sigma_t) - \alpha(I - VV^T)\nabla_{\hat{x}_t} L(D_\theta(\hat{x}_t, \sigma_t)),
\end{equation}
where
\begin{equation}
\label{eq:constraint_L2}
    L(\hat{x}_t) = \|CD_\theta(\hat{x}_t, \sigma_t) - y\|^2
\end{equation}
is a loss function measuring how well the denoiser output conforms to the imposed constraint. This is the post-processed denoiser function \eqref{eq:conditioned_denoiser} in the main text. The extra correction term effectively induces a gradient descent roughly in the form
\begin{equation}
\label{eq:constraint_gd}
    \dot{\hat{x}}_t = - \hat{\alpha}\nabla_{\hat{x}_t} L(D_\theta(\hat{x}_t, \sigma_t))
\end{equation}
with respect to loss function $L$ \textit{in the dynamics of the unconstrained components}. $\hat{\alpha}$ is a positive "learning rate" that is determined empirically such that the loss value reduces adequately close to zero by the conclusion of the denoising process. Besides the $1/s_t\sigma_t^2$ scaling
bestowed by the diffusion process, it also depends on the scaling of the constraint matrix $C$, and in turn directly influences the permissible solver discretization during sampling. Thus it needs to be tuned empirically. 

Substituting $\tilde{D}_{\theta, \text{cons}}$ for $D_\theta(\hat{x}_t, \sigma_t)$ in~\eqref{eq:denoiser_to_score} results in the conditional score
\begin{equation}
\label{eq:constrained_denoiser_to_score}
    \nabla_{x_t}\log{p_t(x_t|E_y)} = \frac{\tilde{D}_{\theta, \text{cons}}(\hat{x}_t, \sigma_t) - \hat{x}_t}{s_t\sigma_t^2}.
\end{equation}
Note that the same re-scale $1/s_t\sigma_t^2$ is applied as before.

\paragraph{Remark 1.} The correction in~\eqref{eq:pp_target_2} is equivalent to imposing a Gaussian likelihood on $x_0$ (and thus the linearly transformed $Cx_0$) given $x_t$. To see this, first note that that applying Bayes' rule to the conditional score function results in
\begin{equation}
    \nabla_{x_t}\log{p_t(x_t|E_y)} = \nabla_{x_t}\log{p_t(x_t)} + \nabla_{x_t}\log{p(Cx_0 = y | x_t)},
\end{equation}
where the probability in the second term may be viewed as a likelihood function for $Cx_0$.

Next, substituting~\eqref{eq:pp_target_2} into ~\eqref{eq:constrained_denoiser_to_score} yields
\begin{equation}
    \nabla_{x_t}\log{p_t(x_t|E_y)} = \frac{D_{\theta, \text{cons}}(\hat{x}_t, \sigma_t) - \hat{x}_t}{s_t\sigma_t^2} + (I - VV^T)\frac{-\alpha}{s_t\sigma_t^2}\nabla_{\hat{x}_t} \|CD_\theta(\hat{x}_t, \sigma_t) - y\|^2
\end{equation}
where the second term (without the projection) may be further rewritten as
\begin{align}
    -\frac{\alpha}{s_t\sigma_t^2}\nabla_{\hat{x}_t} \|CD_\theta(\hat{x}_t, \sigma_t) - y\|^2 &= -\frac{\alpha}{\sigma_t^2}\nabla_{x_t} \|CD_\theta(\hat{x}_t, \sigma_t) - y\|^2 \\
    &= \nabla_{x_t} \log \left(\exp\left(-\frac{2\alpha}{2\sigma_t^2}\|CD_\theta(\hat{x}_t, \sigma_t) - y\|^2\right)\right) \\
    &= \nabla_{x_t} \log \mathcal{N}\left(y; CD_\theta(\hat{x}_t, \sigma_t), \frac{\sigma_t^2}{2\alpha}I\right).
\end{align}
In other words, the correction is equivalent to imposing an isotropic Gaussian likelihood model for $y$ with mean $CD_\theta(\hat{x}_t, \sigma_t)$ and variance $\sigma_t^2/2\alpha$. It is worth noting that both the mean and variance here have direct correspondence to estimations of statistical moments using Tweedie's formulas~\cite{efron2011tweedie}:
\begin{align}
    \mathbb{E}[x_0|x_t] & = D_\theta(\hat{x}_t, \sigma_t) \qquad\text{and} \qquad \text{Cov}[x_0|x_t] = \sigma_t^2\nabla_{\hat{x}_t} D_\theta(\hat{x}_t,t),
\end{align}
with an additional approximation for the (linearly transformed) covariance
\begin{equation}
    C\nabla_{\hat{x}_t} D_\theta(\hat{x}_t,t)C^T \approx \frac{1}{2\alpha}I,
\end{equation}
which is expensive to evaluate in practice.

Lastly, it is important to note that the true likelihood $p(x_0|x_t) \propto p(x_t|x_0)p(x_0)$ is in general not Gaussian unless the target data distribution $p(x_0)$ is itself Gaussian. However, the Gaussian assumption is good at early stages of denoising ($t \gg 0$) when the signal-to-noise ratio (SNR) is low. Later on, the true likelihood becomes closer to a $\delta$-distribution as $\sigma_t\rightarrow 0$, and the denoising is in turn dictated by the mean.

\paragraph{Remark 2.} The post-processing presented in this section is similar to~\cite{chung2022diffusion}, 
who propose to apply a correction proportional to $\nabla_{\hat{x}_t} \|CD_\theta(\hat{x}_t, \sigma_t) - y\|^2$ directly to the score function. The main difference is the lack of the additional scaling $\sigma^2_t$ that adapts to the changing noise levels in the denoise process. In practice, we found that including this scaling contributes greatly to the numerical stability and efficiency of continuous-time sampling.

\section{Diffusion model details}
\label{app:diffusion}

\subsection{Training}
The training of our denoiser-based diffusion models largely follows the methodology proposed in~\cite{karras2022elucidating}. In this section, we present the most relevant components for completeness and better reproducibility.

The variance-preserving (VP) schedule sets the forward SDE parameters:
\begin{equation}
\label{eq:vp}
    \sigma_t = \sqrt{e^{\frac{1}{2}\beta_{\text{d}} t^2 + \beta_{\text{min}}t} - 1}, \qquad s_t = 1/\sqrt{e^{\frac{1}{2}\beta_{\text{d}} t^2 + \beta_{\text{min}}t}} = 1/\sqrt{\sigma_t^2 + 1},
\end{equation}
with $\beta_{\text{b}} = 19.9$, $\beta_{\text{min}} = 0.1$ and time $t$ going from 0 to 1.

The loss function for training the denoiser $D_\theta$ reads as
\begin{equation}
\label{eq:denoiser_loss}
    L(\theta) = \sum_i\lambda(\sigma_{t_i})\|D_\theta(x_{0, i} + \sigma_{t_i}\varepsilon, \sigma_{t_i}) - x_{0, i}\|^2
\end{equation}
over a batch $\{x_{0, i}, t_i\}$ of size $N_{\text{batch}}$ indexed by $i$, with $x_{0, i}\sim p_{\text{data}}$ and $\varepsilon\sim\mathcal{N}(0, I)$. The times $\{t_i\}$ are selected such that
\begin{equation}
\label{eq:train_noise_sampling}
    t_{i+1} - t_i = \Delta t, \quad t_0 \sim \mathcal{U}[\epsilon_t, \epsilon_t + \Delta t], \quad \Delta t = \frac{1 - \epsilon_t}{N_{\text{batch}}}.
\end{equation}
That is, the times are evenly spaced out in $[\epsilon_t, 1]$ with interval $\Delta t$ given a random starting point $t_0$. $\epsilon_t = 10^{-3}$ is the minimum time set to prevent numerical blow-up. $\lambda$ is the weight assigned to the loss at noise level $\sigma_{t_i}$, which is given by
\begin{equation}
\label{eq:train_noise_weight}
    \lambda(\sigma_{t_i}) = (\sigma_{t_i}^2 + \sigma_{\text{data}}^2)/(\sigma_{t_i}\sigma_{\text{data}})^2,
\end{equation}
where $\sigma_{\text{data}}^2$ is the data variance.

We further adopt the preconditioned denoiser ansatz
\begin{equation}
\label{eq:denoiser_precondition}
    D_\theta(x, \sigma) = \frac{\sigma_{\text{data}}^2}{\sigma_{\text{data}}^2 + \sigma^2}x + \frac{\sigma_{\text{data}}\sigma}{\sqrt{\sigma_{\text{data}}^2 + \sigma^2}}F_\theta\left(\frac{x}{\sqrt{\sigma_{\text{data}}^2 + \sigma^2}}, \frac{1}{4}\log(\sigma)\right),
\end{equation}
where $F_\theta$ is the raw U-Net model. This ansatz ensures that the training inputs and targets of $F_\theta$ both roughly have unit variance, and the approximation errors in $F_\theta$ are minimally amplified in $D_\theta$ across all noise levels (see Appendix B.6 in~\cite{karras2022elucidating}).

\paragraph{Data augmentation} Both KS and NS exhibit translation symmetry (only in the $x$-direction for NS due to the $y$-dependence of the Kolmogorov forcing, see section~\ref{app:datasets}), meaning that $u(x + a)$ (or $u(x+a, y)$ for NS) is automatically a valid sample for any constant scalar $a$ provided that $u(x)$ (or $u(x,y)$ for NS) is a valid sample. We leverage this property, as well as the fact that both systems are subject to periodic boundary conditions, to augment our dataset by applying a \texttt{numpy.roll} operation with a random shift.

\subsection{Sampling}

The reverse SDE in~\eqref{eq:bwd_sde} used for sampling may be rewritten in  terms of denoiser $D_\theta$ as
\begin{equation}
\label{eq:bwd_sde_denoiser}
    dx_t = \left[\left(\frac{\dot{\sigma}_t}{\sigma_t} + \frac{2\dot{s}_t}{s_t}\right) x_t - \frac{2s_t\dot{\sigma}_t}{\sigma_t}D_\theta\left(\frac{x_t}{s_t}, \sigma_t\right)\right]dt + s_t\sqrt{2\dot{\sigma}_t\sigma_t}\;dW_t.
\end{equation}
Parts of the drift term inside the squared brackets are inversely proportional to $\sigma_t$ and hence quickly rises in magnitude as $\sigma_t\rightarrow 0$. This means that the dynamics becomes \emph{stiffer} as $t\rightarrow 0$, necessitating the use of progressively finer time steps during denoising. As stated in \S\ref{sec:data} of the main text, for this very reason, we employ an exponential profile with non-uniform time steps proportional to $\sigma_t$.

Similarly, the stiffness of the dynamics also increases with the conditioning strength $\alpha$ in the post-processed denoiser $\tilde{D}_\theta$ in \eqref{eq:conditioned_denoiser}. Therefore, for each conditional sampling setting (downscaling factor and $C'$ map), we use an \textit{ad hoc} number of steps, as determined empirically from a grid search (section \ref{app:ablation_cond_strength}).

\section{Metrics} 
\label{app:metrics}

\subsection{Definitions}
\label{app:metric_definitions}

In this section, we present the definitions of additional metrics that are used for the comparisons in  \S\ref{sec:main_results}. The energy-based metrics are already defined in~\eqref{eq:energy_spectrum} and~\eqref{eq:MELR} of the main text.

\textbf{Relative root mean squared error (RMSE) }
is defined as
\begin{equation}
\label{eq:RMSE}
    \text{RMSE} = \frac{1}{N}\sum_{n=1}^N \frac{\|z_{\text{pred}, n} - z_{\text{ref}, n}\|_2}{\|z_{\text{pred}, n}\|_2},
\end{equation}
where the predicted and reference quantities $z_{\text{pred}, n}$ and $z_{\text{ref}, n}$ are computed over an evaluation batch (of size $N$, indexed by $n$). The \emph{constraint RMSE} corresponds to this metric evaluated on the conditioned pixels of the generated samples (predicted) and the conditioned values $\bar{y}'$ (reference). It provides a measure for how well the generated conditional samples satisfy the imposed constraint.

\textbf{Covariance RMSE (covRMSE)}
referenced in Table~\ref{table:OT_results} corresponds to computing \eqref{eq:RMSE} between the (empirical) covariance matrices of the generated and reference samples given by
\begin{equation}
\label{eq:empirical_covariance}
    \text{Cov}(u) = \frac{1}{N}\sum_{n=1}^N (u_n - \bar{u})(u_n - \bar{u})^T, \quad \bar{u} = \frac{1}{N}\sum_{n=1}^N u_n,
\end{equation}
where $u_n$ are realizations of the multi-dimensional random variable $U$.
For KS, we compute the covariance along the full domain by treating each pixel as a distinct dimension of the random variable. For NS, we leverage the translation invariance in the system to compute the covariance on slices with fixed $x$-coordinate (i.e., dimensions are indexed by the $y$-coordinate). Lastly, since we are dealing with matrices, the norm involved in \eqref{eq:RMSE} is taken to be the Frobenious norm. 

\textbf{Kernel-density-estimated Kullback-Leibler divergence (KLD) }
computes the KL divergence using 1-dimensional marginal kernel density estimations (KDEs, with the bandwidths selected based on Scott's rule~\cite{scott2015multivariate}). That is,
\begin{equation}
\label{eq:KDE_KLD}
    \text{KLD} = \sum_{m=1}^d \int_{-\infty}^{\infty} \tilde{p}_{d, \text{ref}}(\upsilon)\log\left(\frac{\tilde{p}_{d, \text{ref}}(\upsilon)}{\tilde{p}_{d, \text{pred}}(\upsilon)}\right) d\upsilon,
\end{equation}
where $\tilde{p}_m$ are empirical probability density functions (PDFs) obtained with KDE for a particular dimension $m$ of the samples. The integral is approximated using the trapezoidal rule, and summed over all dimensions for an aggregated measure, as if they were independent.

\textbf{Sample variability (Var)} refers to the mean pixel-wise standard deviation in the generated conditional samples given by
\begin{equation}
\label{eq:variability}
    \text{Var} = \sqrt{\frac{1}{Nd}\sum_n^N\sum_m^d (u_{nm} - \bar{u}_m)^2}, \quad \bar{u}_m  = \frac{1}{N}\sum_{n=1}^N u_{nm},
\end{equation}
where $\bar{u}_m$ is obtained by averaging the values of dimension $m$ over samples \emph{with the same condition}.

\textbf{Mean Maximum Discrepancy (MMD)} is computed using the following empirical estimation
\begin{equation}
\label{eq:mmd}
\begin{aligned}
    \text{MMD}^2 = &\frac{1}{N_p(N_p-1)}\sum_{i, j\ne i}k(z_{\text{pred}, i}, z_{\text{pred}, j}) - \frac{2}{N_pN_r}\sum_{i, j}k(z_{\text{pred}, i}, z_{\text{ref}, j}) \\ &+ \frac{1}{N_r(N_r-1)}\sum_{i, j\ne i}k(z_{\text{ref}, i}, z_{\text{ref}, j}),
\end{aligned}
\end{equation}
between generated and reference samples $\{z_\text{pred}\}$ and $\{z_\text{ref}\}$. For $k$ we use a multi-scale Gaussian kernel with bandwidths $[2, 4, 6, 8] \times 256$, which are tuned empirically to the rough scales of the reference distribution.

\textbf{Wasserstein-1 metric (Wass1)} is given by
\begin{equation}
\label{eq:wass1}
    \text{Wass1} = \frac{1}{d}\sum_m^d \int\big|\text{CDF}_{\text{pred}, d}(z) - \text{CDF}_{\text{ref}, d}(z)\big|\; dz,
\end{equation}
where the 1-dimensional CDFs are empirically computed with \texttt{np.histogram} and averaged across all dimensions. The integral is performed over the range $[-20, 20]$.

\paragraph{Evaluation setup.} The number of samples used to evaluate the metrics is summarized in Table~\ref{table:evaluation_setup}. MELR and KLD metrics are evaluated marginally, i.e., on all conditional samples pooled together. 

\begin{table}[t]
\centering
\caption{Number of samples used for evaluation. For sampling runs which are deterministic or unconditional in nature, the number of evaluation samples is equal to the number of OT samples (rather than the number of conditions) to ensure convergence in statistics.}

\vspace{2pt}
{\setlength\tabcolsep{5pt} {\setlength{\extrarowheight}{2pt}
\begin{tabular}{l|c|cc|}
\cline{2-4}
                                      & \textbf{OT} & \multicolumn{2}{c|}{\textbf{Conditional sampling}} \\ \hline
\multicolumn{1}{|l|}{\textbf{System}} & samples     & conditions         & samples per condition         \\ \hline
\multicolumn{1}{|l|}{KS}              & 15360       & 512                & 128                           \\
\multicolumn{1}{|l|}{NS}              & 10240       & 128                & 128                           \\ \hline
\end{tabular}
}}
\label{table:evaluation_setup}
\end{table}

\begin{table}[t]
\centering
\caption{Additional KS conditional sampling metrics.}

\vspace{2pt}
{\setlength\tabcolsep{5pt} {\setlength{\extrarowheight}{2pt}
\begin{tabular}{|l|ccccc|}
\hline
\textbf{Method}   & \begin{tabular}[c]{@{}c@{}}Constraint \\ RMSE\end{tabular} & \begin{tabular}[c]{@{}c@{}}Sample\\ Variability\end{tabular} & \begin{tabular}[c]{@{}c@{}}MELR\\ (unweighted)\end{tabular} & \begin{tabular}[c]{@{}c@{}}MELR\\ (weighted)\end{tabular} & KLD \\ \hline
\textit{Raw cDfn} & 0.001                                                      & 0.044                                                        & 0.527                                                       & 0.143                                                     & 10.37   \\
\textit{OT+cDfn}  & 0.001                                                      & 0.044                                                        & 0.362                                                       & 0.044                                                     & 1.27    \\ \hline
\end{tabular}
}}
\label{table:KS_additional_metrics}
\end{table}

\begin{table}[t]
\centering
\caption{LR metrics for NS, computed for downsampled outputs of end-to-end baselines (BCSD, cycGAN and ClimAlign). OT is superior in distributional metrics (covRMSE, MELR, MMD) but pays a "price" in terms of pixel-wise similarity represented in the sMAPE metric (between corrected and uncorrected LR snapshots).}

\vspace{2pt}
{\setlength\tabcolsep{5pt} {\setlength{\extrarowheight}{2pt}
\begin{tabular}{lcccc}
\hline
             & OT   & BCSD & cycGAN & ClimAlign \\ \hline
\textbf{8$\times$downscale}  &      &      &        &           \\ \hline
covRMSE          & 0.08 & 0.31 & 0.16   & 2.21      \\ \hline
MELRu        & 0.01 & 0.95 & 0.08   & 0.53      \\ \hline
MELRw        & 0.03 & 0.13 & 0.04   & 0.54      \\ \hline
MMD          & 0.04 & 0.06 & 0.06   & 0.61      \\ \hline
sMAPE        & 0.53 & 0.25 & 0.41   & 0.74      \\ \hline
\textbf{16$\times$downscale} &      &      &        &           \\ \hline
covRMSE          & 0.08 & 0.35 & 0.33   & 2.50      \\ \hline
MELRu        & 0.02 & 0.63 & 0.34   & 0.67      \\ \hline
MELRw        & 0.03 & 0.16 & 0.15   & 0.58      \\ \hline
MMD          & 0.03 & 0.34 & 0.09   & 0.55      \\ \hline
sMAPE        & 0.54 & 0.36 & 0.63   & 0.76      \\ \hline
\end{tabular}
}}
\label{table:NS_additional_lowres_metrics}
\end{table}

\begin{figure}[t]
  \centering
  {\setlength\tabcolsep{0.5pt} {\renewcommand{\arraystretch}{0.5}
      \begin{tabular}{M{.32\linewidth}M{.32\linewidth}M{.32\linewidth}}
        \hspace{-3pt}\includegraphics[width=\linewidth]{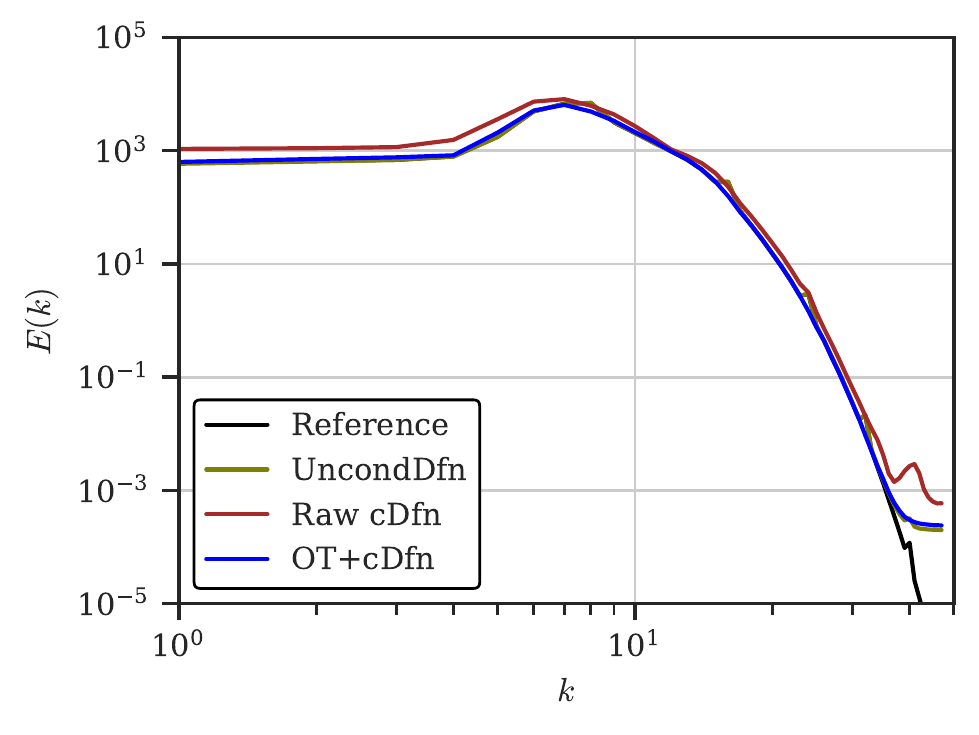} &
        \hspace{-3pt}\includegraphics[width=\linewidth]{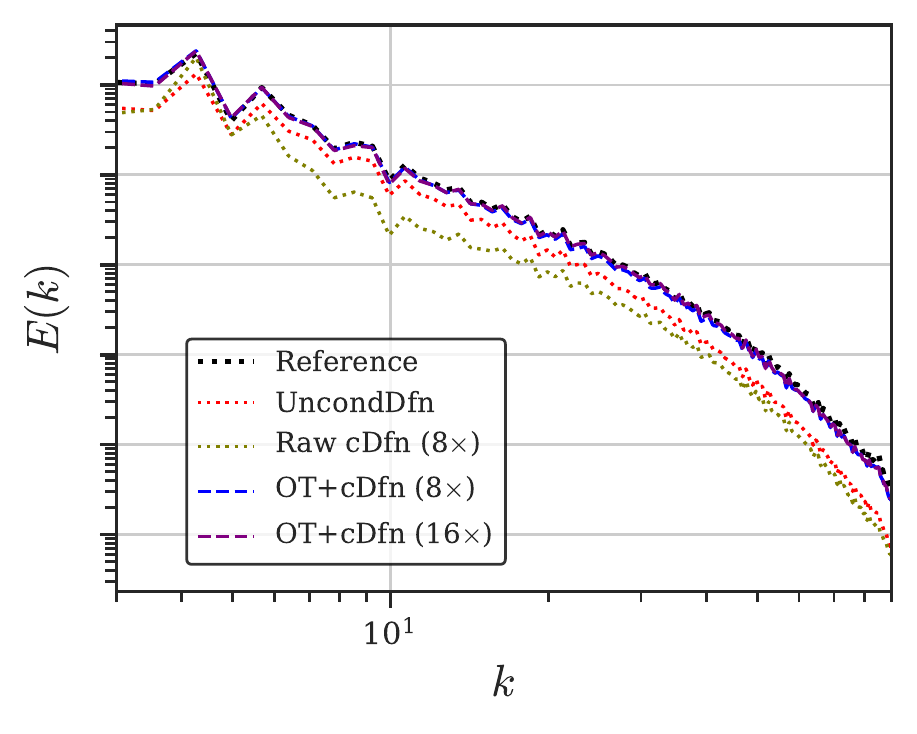} & 
        \hspace{-3pt}\includegraphics[width=\linewidth]{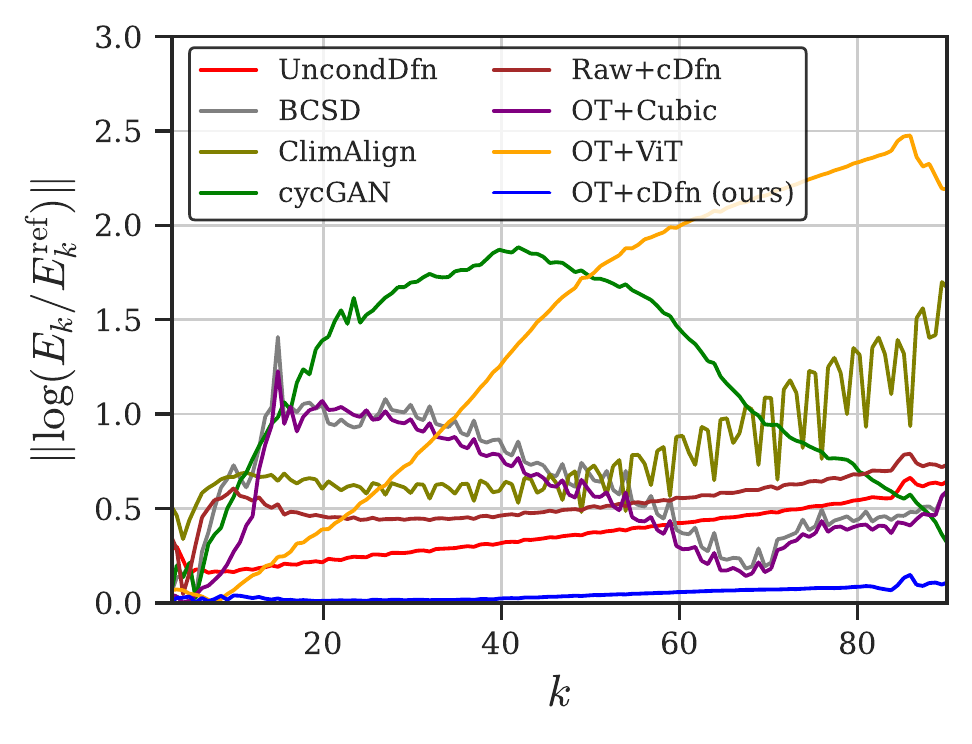} \\
        (a) KS, energy spectra & \quad (b) NS, energy spectra & \quad (c) NS $16\times$, log energy ratios \\
      \end{tabular}
  }}
  \caption{(a) Sample energy spectra (\eqref{eq:energy_spectrum}) comparison in KS, (b) sample energy spectra comparison in NS and (c) mode-wise log energy ratios with respect to the true samples (\eqref{eq:MELR} without weighted sum) at $16\times$ downscaling for NS.}
  \label{fig:additional_metrics}
\end{figure}

\subsection{Additional results}

Table~\ref{table:KS_additional_metrics} shows the conditional sampling metrics for KS. Additional energy spectra and log energy ratio calculations for both systems are displayed in Fig.~\ref{fig:additional_metrics}.

Table~\ref{table:NS_additional_lowres_metrics} contrasts OT with end-to-end baselines. Since end-to-end baselines directly output HR samples, they are downsampled to LR to enable apple-to-apple comparison. OT achieves the best distributional metrics. Note that this comes seemingly "at the price" of decreased pixel-wise similarity, which may be quantified through the symmetric mean absolute percentage error (sMAPE):
\begin{equation}
    \text{sMAPE} = \frac{1}{N}\sum_{n=1}^N \frac{|y_n - y_n'|}{(|y_n| + |y_n'|) / 2},
    \label{eq:sMAPE}
\end{equation}
where $y_n$ and $y'_n$ denote the LFLR and the downsampled end-to-end baseline outputs. Note that sMAPE more closely embodies the "visual discrepancy" one observes before and after debiasing. We reemphasize that this is a feature inherent to distribution-based debiasing and in fact an intended consequence.

\section{Baselines}
\label{app:baselines}

\subsection{Cubic interpolation}

Cubic interpolation employs a local third-order polynomial for the interpolation process. It builds a local third order polynomial, or a cubic spline, in the form 
\begin{equation}
    u(x, y) = \sum_{i=0}^3 \sum_{j=0}^3 a_{ij} x^i y^j, 
\end{equation}
where the coefficients $a_{ij}$ are usually found using Lagrange polynomials. We use the function \texttt{jax.image.resize} to perform the interpolation. 

\subsection{BCSD}

Bias correction and statistical disaggregation (BCSD) \cite{Maraun2013:quantile_matching} is a two-stage downscaling procedure. It first implements a cubic interpolation (using \texttt{jax.image.resize}), and then performs pixel-wise quantile matching.

For quantile matching, we use the \texttt{tensorflow.probability} library. Specifically, for each point in both the interpolated and reference HRHF snapshots, we compute the segments corresponding to 1000 quantiles of each distribution using the \texttt{stats.quantiles()} function. At inference time, for each pixel in the interpolated snapshot, we perform the following steps: (i) find the closest segment (out of the 1000 quantiles) that contains the value at the given pixel, (ii) identify the quantile corresponding to that segment, (iii) find the segment corresponding to that quantile in the HRHF data, and finally,
(iv) output the middle point in that quantile.

The number of quantiles was chosen to minimize the Wasserstein norm, while reliably computing the quantiles from the $90,000$ samples. Finally, we note that quantile matching is indeed the minimizer of the Wasserstein-1 norm, which in the one dimensional case can be conveniently expressed as the $L^1$ distance between the cumulative distribution functions of the corresponding measures.

\subsection{ViT model}

For the deterministic upsampling model, we consider a Vision Transformer (ViT) model similar to several CNN based super-resolution models \cite{dong2014learning}, but with a Transformer core that significantly increases the capacity of the model. Our model follows the standard structure of a ViT. However, it differs in the tokenization step, where instead of using a linear transformation from a patch of the input to a embedding, we employ a single-pixel embedding combined with a series of downsampling blocks. Each downsampling block consists of a sequence of ResNet blocks and a coarsening layer implemented using a strided convolution. This architectural choice draws inspiration from the hierarchical processing of CNNs~\cite{dong2014learning,Zhu2017:CycleGAN}. After tokenization, the tokens are processed using self-attention blocks following \cite{dosovitskiy2021an}. The outputs of the self-attention blocks are then upsampled using upsampling blocks. Each upsampling block consists of a nearest neighbor upsampling layer, which combines nearest neighbor interpolation and a convolution layer, followed by a sequence of ResNet blocks. We provide below more details on the implementation of each block and the core architecture. 

\textbf{Embedding } 
We use a $1\times 1$ convolution to implement a pixel-wise embedding whose dimension was tuned in a hyperparameter sweep.

\begin{figure}
\centering
\includegraphics[scale=0.90, trim={0cm 11cm 10cm 0cm}, clip]{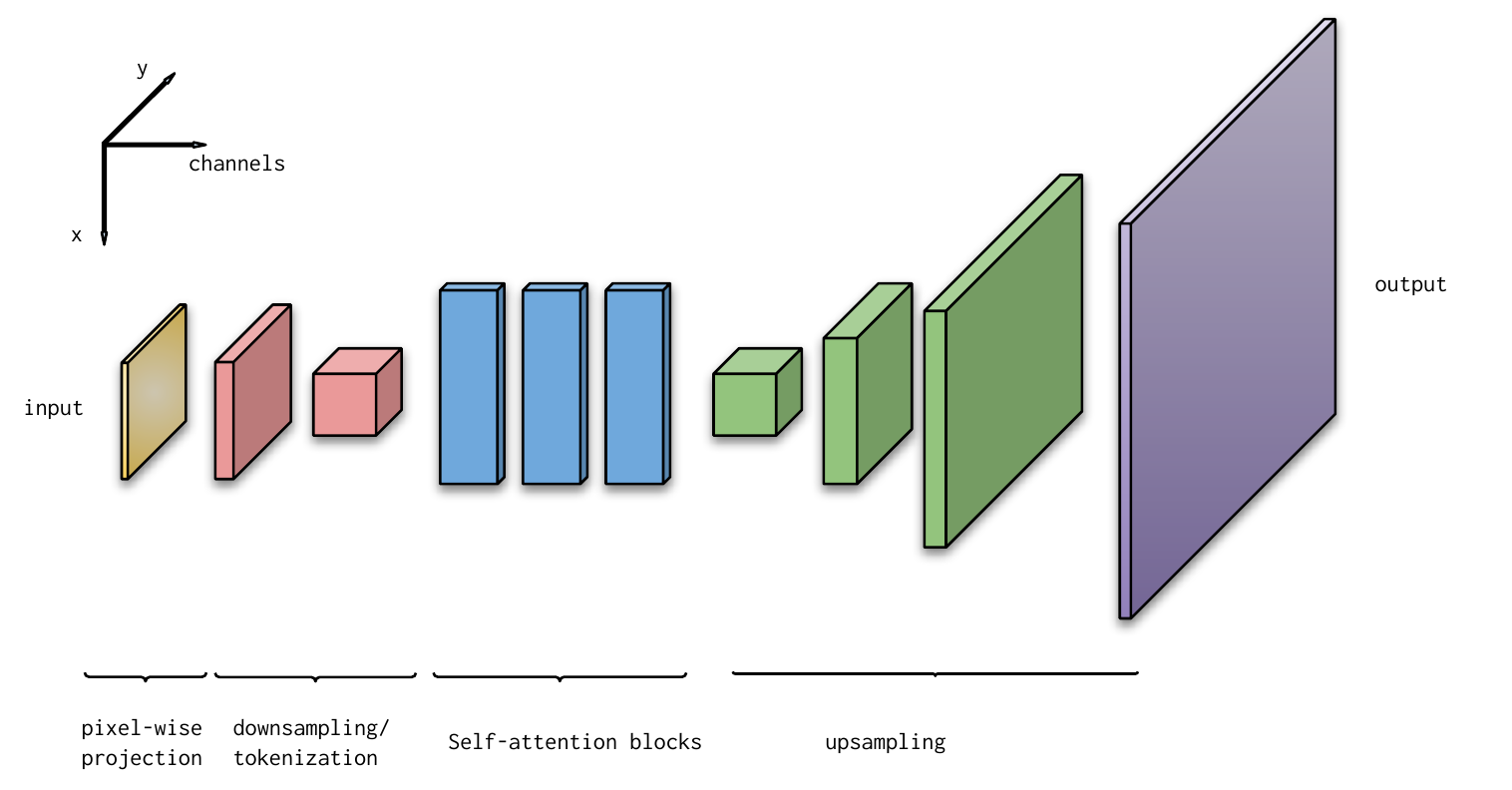}
\caption{Sketch of the structure of the deterministic upsampling network. In red we have the downsampling layers, which reduce the spatial resolution in space of the image, while increasing the number of channels. The first two channels are flattened into a one-dimensional set of tokens, and a sequence of self-attention blocks are applied afterwards. After the self-attention blocks, the tokens are reshaped back to it two-dimensional geometry, followed by a cascade of transpose convolutions to upsample the image.}
\label{fig:ViT}
\end{figure}

\textbf{Downsampling blocks } 
The downscaling blocks quadruple the number of channels of the input as the other dimensions are decimated by a factor two (red blocks in Fig. \ref{fig:ViT}). This is achieved with a convolutional layer with a $(2,2)$ stride, a fixed kernel width (hyperparameter) and periodic (i.e. circular) boundary conditions. After the convolution, we use a sequence of convolutional ResNet layers, with a GeLU activation function and a layer normalization. These convolutional layers also use a fixed kernel width and periodic boundary conditions. The number of downsampling blocks and the number of ResNet blocks inside each layer were empirically tuned.

\textbf{Transformer core } 
Then the output of the downsampling blocks is reshaped into a sequence of tokens, in which a corresponding 2-dimensional embedding is added to account for the underlying geometry in the self-attention blocks. We then perform a sequence of self-attention blocks following the blocks introduced in \cite{dosovitskiy2021an} (blue blocks in Fig. \ref{fig:ViT}). The number of attention heads of each self-attention block is equal to the dimension. The number of transformer blocks is also tuned. The GeLU activation functions is used for the self-attention layers.  

\textbf{Upsampling blocks } 
The tokens are reshaped back to their original 2-dimensional topology. Then, a sequence of upsampling layers followed by a handful of ResNet blocks is used to downscale the image (green blocks in Fig. \ref{fig:ViT}) at its target resolution (purple block in Fig. \ref{fig:ViT}). At each upsampling step, the spatial resolution is increased by a factor of two in each dimension, while reducing the number of channel so that the overall information remains constant. As mentioned above, the upsampling is performed using a nearest neighbor interpolation (we repeat the value of the closest neighbor), followed by a linear convolutional layer with a $(3,3)$ kernel size, and subsequently a series of ResNet blocks similar to those used in the downsampling block. 

The number of downsampling and upsampling blocks were chosen to strike a computational balance between the quadratic complexity of the Transformer core on the number of tokens and the quadratic complexity on the width of each token. 

We use a simple mean squared error loss given that in this case we have paired data. As we seek to build a upsampling network from $\mathcal{Y}'$ to $\mathcal{X}$, the inputs $y' \in \mathcal{Y}'$ are nothing more than the desired output but downsampled by a factor $8$ or $16$, or $y' = C' x$ for $x \in \mathcal{X}$. We observed that the network in the first runs were not able to faithfully interpolate the input, i.e., if we denote the network by $\mathcal{N}_{\theta}$, where $\theta$ corresponds to the set of parameters, then the interpolation should satisfy $ C' \mathcal{N}_{\theta} (y') = (y')$. Therefore, we added a regularization term in the loss weighted by a tunable parameter $\lambda$. In a nutshell, the loss is given by
\begin{equation}
    \mathcal{L}(\theta) = \frac{1}{N} \left ( \sum_{ x \in \mathcal{X}}  \| \mathcal{N}_{\theta}( C'x) - x \|^2  +  \lambda \| C'\mathcal{N}_{\theta}( C'x) - C'x \|^2  \right).
\end{equation}

For training, we used a regular \texttt{adam} optimizer. Due to the Transformer core we used a small learning rate of $3\cdot 10^{-3}$ with a gentle decay of $0.97$ every $40,000$ iterations, and a batch size of $32$. We trained the network for $800,000$ iterations using the same data used to train our models. 
We performed hyperparameter sweeps on the embedding dimension, the number of ResNet blocks, the number of self-attention layers, kernel sizes and regularization parameter $\lambda$. We observed that increasing $\lambda$ did not provide much overall performance boost, so the final version was trained with $\lambda$ equal to zero. Also, adding more blocks and self-attention layers saturated the performance quickly. For each combination of hyperparameters, the training took between $14$ and $27$ hours depending on the number of trainable parameters.
The models used for the comparison in Table~\ref{table:ns_metrics} have the following hyperparameters: 
\begin{itemize}
    \item \textbf{$8\times$ downscaling}: dimension embedding in the embedding layer - $16$; number of downsampling blocks - $2$; number of upsampling blocks - 5; number of ResNet block (in each of the up-sampling/downsampling) - $4$; number of self-attention blocks - $2$. Total number of parameters: $9,726,209$.

    \item \textbf{$16\times$ downscaling}: dimension embedding in the embedding layer - $32$; number of downsampling blocks - $1$; number of upsampling blocks - $5$; number of ResNet block (in each of the up-sampling/downsampling) - $4$; number of self-attention blocks - $2$. Total number of parameters: $2,669,505$.
\end{itemize}
We point out that the network for the example with $8\times$ downscaling factor has more parameters due to the extra downsampling block which quadruples the number of embedding dimension. We also considered bigger dimension embedding, but we found no considerable gains in performance. 

\subsection{cycleGAN}

For the cycleGAN we followed closely the implementation in \cite{Zhu2017:CycleGAN}. 

For the generators, we use the same architecture as in the original paper. The first layer performs a local embedding, followed by a sequence of downsample blocks, each of which downsamples the geometrical dimensions by a factor two, while increasing the channel dimension. At the lowest resolution we implement a sequence of ResNet blocks to process the input, immediately followed by a sequence of upsampling blocks, which upsample the geometrical dimension while reducing the channel dimension. 

Given that the two generators, $\mathcal{G}_{\mathcal{X}\mapsto \mathcal{Y}}$ (from high-resolution to low-resolution) and $\mathcal{G}_{\mathcal{Y}\mapsto \mathcal{X}}$ (from low-resolution to high-resolution) have different input dimensions, we use a different combination of downsample/upsample blocks, and they also have different embedding dimensions.
We implemented the different generators (for both the $8\times$ and $16\times$ downscaling factor) with different number of downsampling versus upsampling layers in the generators, and also different embedding dimensions. 

Instead of using one discriminator architecture as in the original paper, we use two of them given that the input dimensions are different. Below we provide further details on the architecture used. 

\subsubsection{Generator networks}
\begin{figure}[t]
\centering
\includegraphics[scale=0.85, trim={0cm 11cm 9cm 0cm}, clip]{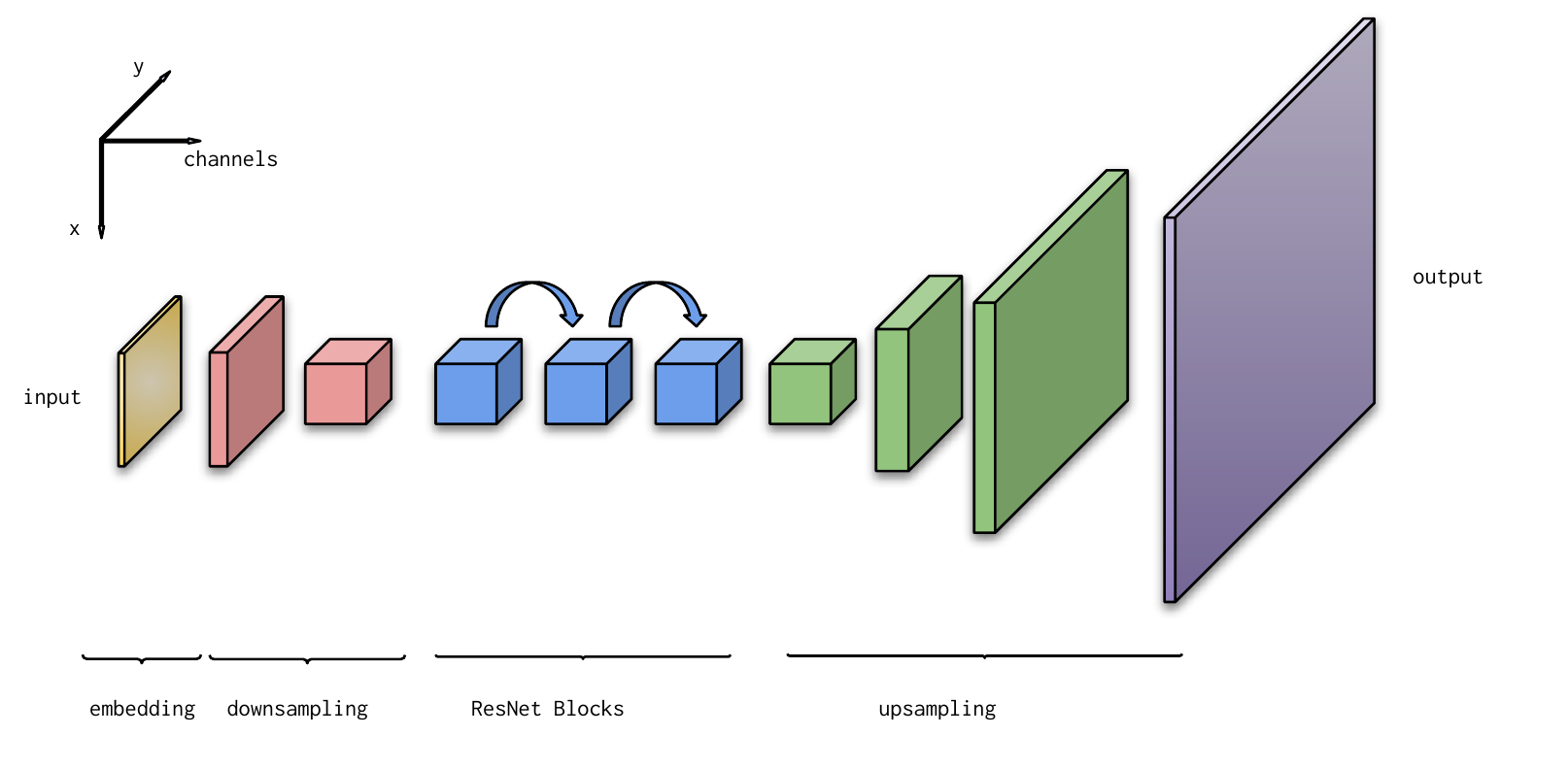}
\caption{Sketch of the structure of the cycleGAN generator $\mathcal{G}_{\mathcal{Y} \mapsto \mathcal{X}}$. In red we have the downsampling layers, which reduce the spatial resolution in space of the image, while increasing the number of channels. At the lowest level we have (in blue) the ResNet blocks, followed by a cascade of transpose convolutions to upsample the image.}
\label{fig:generator_cGAN}
\end{figure}

\textbf{Embedding } 
We use one convolution layer with a kernel of size $(7, 7)$ and an embedding dimension that is different for each generator and for each problem. 

\textbf{Downsampling blocks } 
We implement the downscaling blocks following \cite{Zhu2017:CycleGAN}. These blocks effectively double the number of input channels while reducing the other dimensions by a factor of two. This downsampling is achieved using a convolutional layer with a $(2,2)$ stride, a kernel of fixed width, and periodic boundary conditions. Subsequently, the output is normalized using a group normalization layer and further processed with a ReLU activation function.

\textbf{ResNet core } 
At the lowest resolution we use a sequence (whose length was also tuned) of ResNet blocks, using two convolution layers, with periodic boundary conditions, including a skip connection, two group normalization layers, and a dropout layer with a tunable dropout rate, following \cite{Zhu2017:CycleGAN}.

\textbf{Upsampling blocks } 
The upsampling blocks are implemented using transpose convolutional layers, followed by a group normalization layer and a ReLU activation function.

After several sweeps on the number of upsampling/downsampling blocks and other hyper-parameters we chose the following network configurations. 

\textbf{$8\times$ downscaling factor}
\begin{itemize}
    \item $\mathcal{G}_{\mathcal{Y}\mapsto \mathcal{X}}$: number of downsampling blocks - $2$; number of upsampling blocks - $5$; embedding dimension - $32$; dropout rate - $0.5$; number of ResNet blocks - $6$. Number of parameters: $4,029,569$.

    \item $\mathcal{G}_{\mathcal{X}\mapsto \mathcal{Y}}$: number of downsampling blocks - $5$; number of upsampling blocks - $2$; embedding dimension - $6$; dropout rate - $0.5$; number of ResNet blocks - $6$. Number of parameters: $4,232,353$.
\end{itemize}

\textbf{$16\times$ downscaling factor} 

\begin{itemize}
    \item $\mathcal{G}_{\mathcal{Y}\mapsto \mathcal{X}}$: number of downsampling blocks - $2$; number of upsampling blocks - $6$; embedding dimension - $64$; dropout rate - $0.5$; number of ResNet blocks - $6$. Number of parameters: $16,868,641$, 

    \item $\mathcal{G}_{\mathcal{X}\mapsto \mathcal{Y}}$: number of downsampling blocks - $5$; number of upsampling blocks - $2$; embedding dimension - $6$; dropout rate - $0.5$; number of ResNet blocks - $6$. Number of parameters: $4,232,353$.
\end{itemize}
We point out that in this case the network for the $16\times$ example also requires more parameters: roughly four times more due to the higher dimension of the upsampling. 

\subsubsection{Discriminator networks}

The discriminator networks are the same as those in the original cycleGAN, with a small difference. The discriminator for $\mathcal{X}$ requires a special structure: instead of discriminating the full snapshot, we discriminate patches of the snapshot. By employing this trick, we were able to efficiently train the network, whereas using a global discriminator did not allow us to train the network to generate the snapshots as shown in \S\ref{sec:main_results}. One simple strategy to implement this patched discriminator was to use the same architecture for both discriminators. However, we output a tensor of scores in which each element of the tensor corresponded to the score of one of the patches in the image, rather than a single score for the entire image. By choosing the patch size to be equal to the size of the lowest resolution snapshot, we could reuse the same architecture, depending on the problem size and downscaling factor. 

The discriminator network, as described in \cite{Zhu2017:CycleGAN}, is composed of the following components: an embedding layer that applied a convolution with a kernel size of $(4, 4)$, a stride of two, padding of one, and a tunable embedding dimension; a leaky ReLU applied with an initial negative slope of $0.2$; and a sequence of downsampling blocks similar to the generator network. Finally, a per-channel bottleneck network with one output channel is used to produce the local score.

The specific architectures used in Table~\ref{table:ns_metrics} with their corresponding hyperparameters are summarized below: 

\begin{itemize}
    \item \textbf{$8\times$ downscaling factor.} The discriminators were the same: they had $3$ downsampling blocks, with an embedding dimension of $64$. Total number of parameters $2,763,589$ each.

    \item \textbf{$16\times$  downscaling factor.} The discriminators were different due to the smaller dimensions of the snapshots in $\mathcal{Y}$, which would have resulted in very small receptive fields for the discriminator of $\mathcal{X}$. The discriminator for $\mathcal{Y}$ had two downsampling blocks, while the discriminator for $\mathcal{X}$ had six downsampling blocks. However, both discriminators had an embedding dimension of $64$. The total number of parameters was $15,349,576$ for the discriminator of $\mathcal{X}$ and $661,316$ for the discriminator of $\mathcal{Y}$.
\end{itemize}

\subsubsection{Loss and optimization}

We also closely followed the original cycleGAN paper \cite{Zhu2017:CycleGAN}, in which we utilize the least-squares GAN loss in conjunction with the cycle loss. However, we do not employ the identity loss, as the different dimensions of the spaces make it challenging to impose such a loss naturally.

The optimization was performed by alternating the update of the generators and the discriminators. We used two \texttt{adam} optimizers: one for the generators and the other for the discriminators. Both optimizers had a momentum parameter $\beta$ set to 0.5 and a learning rate of 0.0002. Despite the continuous decrease in losses, we observed the emergence of several artifacts in the generated images. To address this issue, we checkpointed the model every two epochs, computed the MELR (see Equation \eqref{eq:MELR}), and selected the model with the smallest unweighted error. For the example with an $8\times$ downscaling factor, this was achieved after just $8$ epochs, while for the example with a $16\times$ downscaling factor, this was achieved after $16$ epochs.

The full training loop took around two days to complete. However, due to early stopping, the checkpoints shown in Table~\ref{table:ns_metrics} took around $8$ hours to produce.

\subsection{ClimAlign}

For this baseline we follow the original paper ClimAlign \cite{climalign:2021}, in which the authors perform first an cubic interpolation and then use the AlignFlow framework to perform the debiasing.

For the implementation of the debiasing step we follow closely the implementation of the original AlignFlow \cite{grover2019alignflow} algorithm, which can be found in \url{https://github.com/ermongroup/alignflow}. We considered the same hyper parameters as in the original paper. The main modification we perform to the codebase was how to feed the data to the model.

\section{Ablation studies}
\label{app:ablation}

\subsection{Conditioning strengths}
\label{app:ablation_cond_strength}

As described in section~\ref{app:conditioning}, the parameter $\alpha$ controls the strength of conditioning in the subspace orthogonal to the linear constraint. Increasing its value encourages these orthogonal components to be more coherent with the constrained components, but at the same time makes sampling more costly. As such, we conduct grid searches to determine its value, along with the number of SDE solver steps, that strikes a satisfying balance between sample quality and cost.

The grid search setup is as follows: we first normalize $\alpha$ with respect to the dimensionality of $C'$
\begin{equation}
\label{eq:alpha_normalization}
    \tilde{\alpha} = \alpha/\gamma_{C'}, \quad \gamma_{C'} = \text{dim}(\tilde{y}) / \text{dim}(x)
\end{equation}
such that the same normalized $\tilde{\alpha}$ value does not have drastically different effects for different downscaling factors. Then we evaluate the unweighted MELR and sample variability for 2500 generated samples (50 conditions, 50 conditional samples each) resulting from combinations of $\tilde{\alpha}\in[0.125, 0.25, 0.375, ..., 3]$ and $N\in[32, 64, 128, ..., 1024]$.

\begin{figure}[t]
  \centering
  {\setlength\tabcolsep{0.5pt} {\renewcommand{\arraystretch}{0.5}
      \begin{tabular}{M{.32\linewidth}M{.32\linewidth}M{.32\linewidth}}
        \hspace{-3pt}\includegraphics[width=\linewidth]{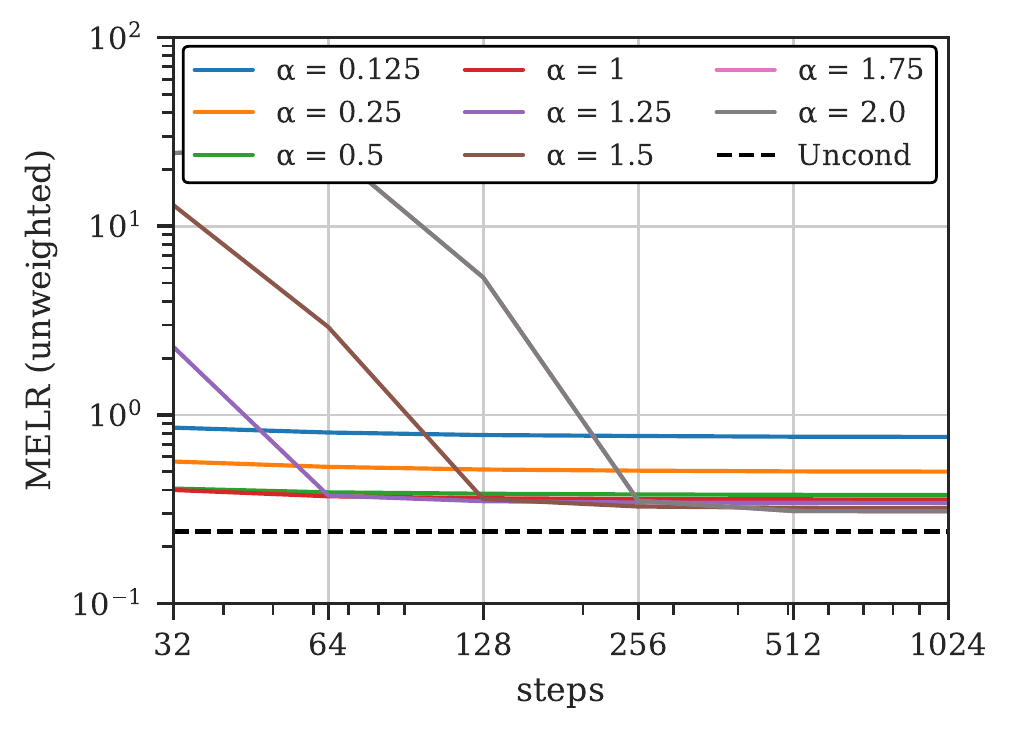} &
        \hspace{-3pt}\includegraphics[width=\linewidth]{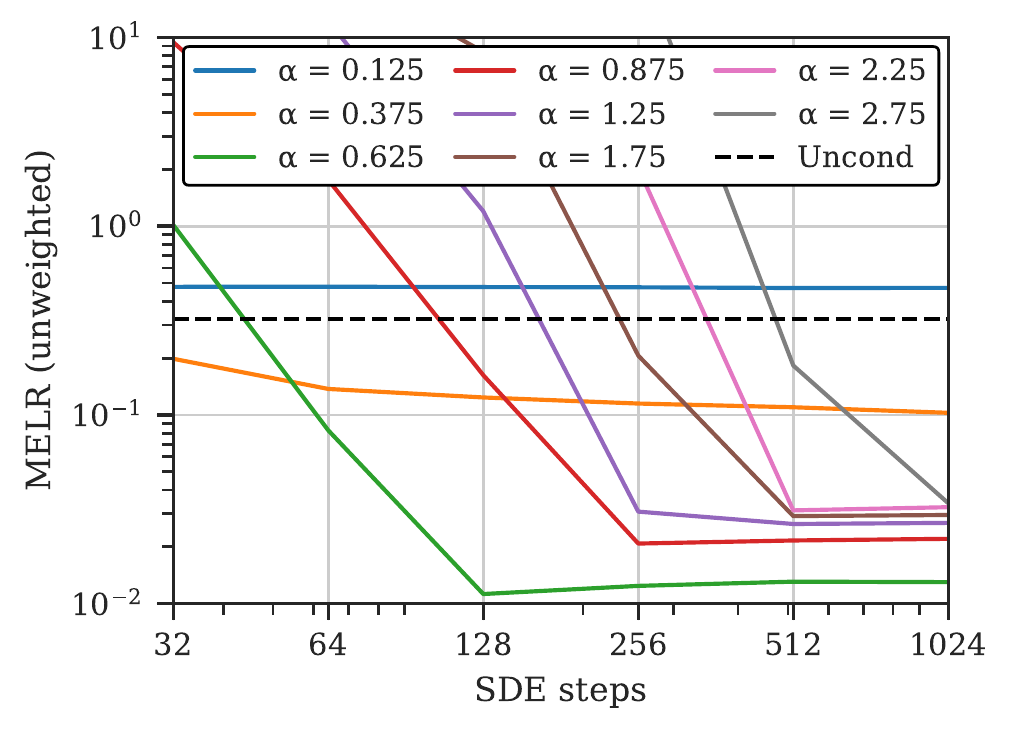} & 
        \hspace{-3pt}\includegraphics[width=\linewidth]{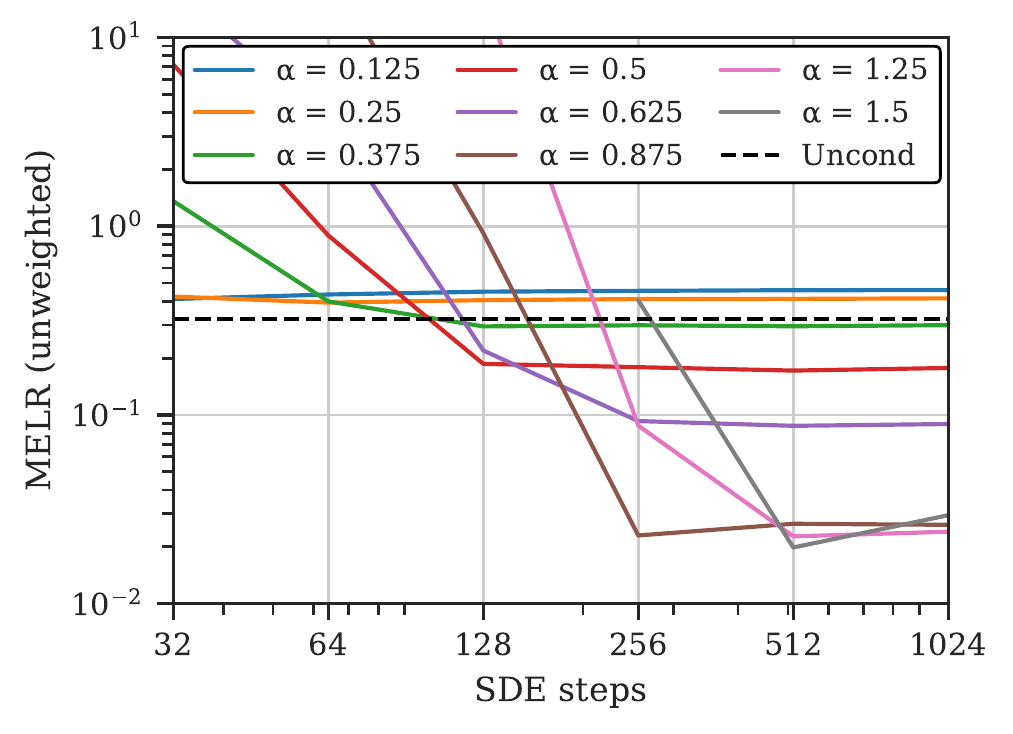} \\
        \hspace{-3pt}\includegraphics[width=\linewidth]{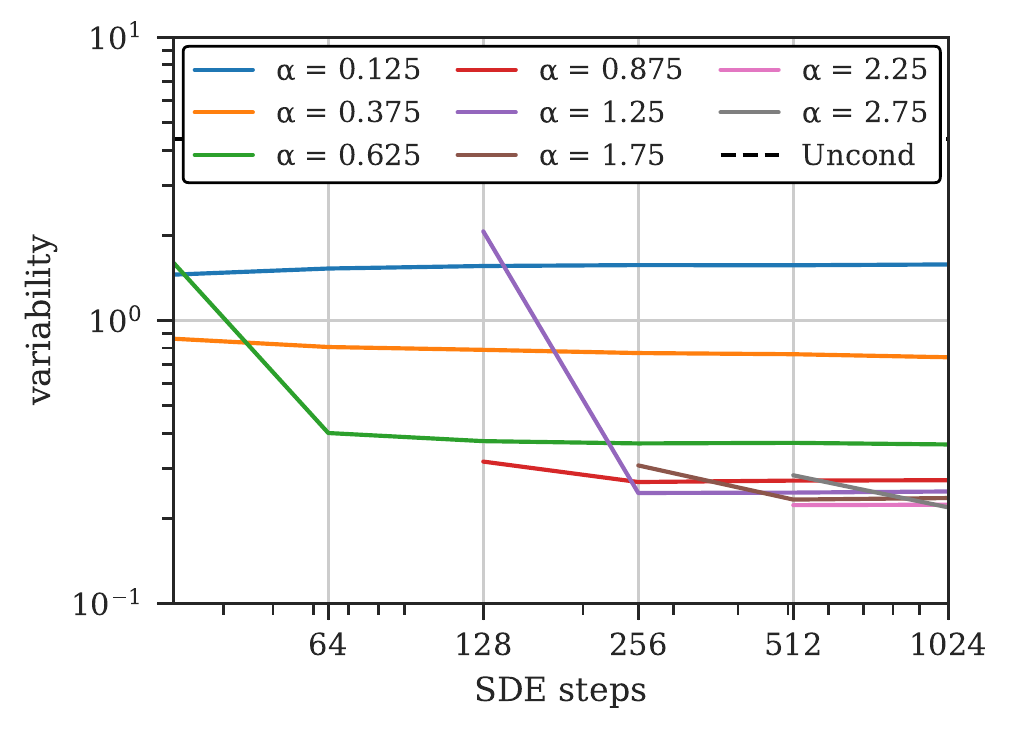} &
        \hspace{-3pt}\includegraphics[width=\linewidth]{figures/ns_std_alpha_steps_8xds.pdf} & 
        \hspace{-3pt}\includegraphics[width=\linewidth]{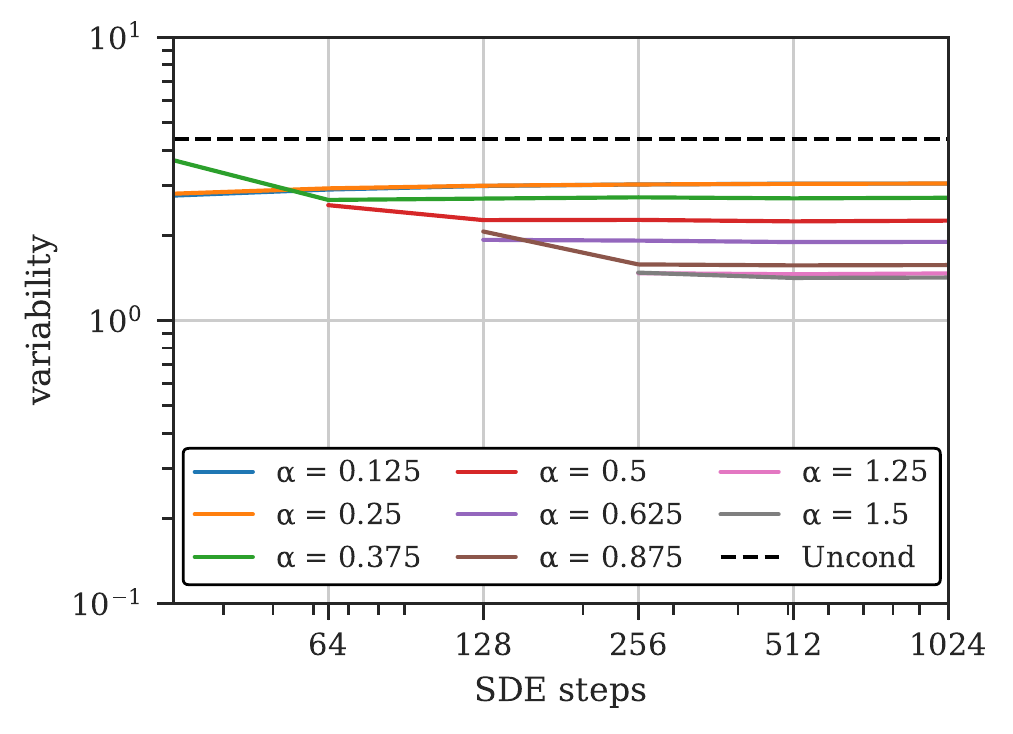} \\
        (a) KS & \quad (b) NS $8\times$ DS & \quad (c) NS $16\times$ DS \\
      \end{tabular}
  }}
  \caption{Unweighted MELR (\eqref{eq:MELR}; first row) and sample variability (\eqref{eq:variability}; second row) vs. number of SDE steps at different values of $\tilde{\alpha}$. Larger $\tilde{\alpha}$ generally has better MELR but takes more solver steps to converge and results in lower sample variability.}
  \label{fig:cond_ablation_melr_variability}
\end{figure}

In Fig.~\ref{fig:cond_ablation_melr_variability}, we show MELR and variability plotted against $N$ for different $\tilde{\alpha}$'s. The MELR trends (first row) confirm our intuition that more steps are required for convergence as $\tilde{\alpha}$ increases. However, higher $\tilde{\alpha}$ also means lower sample variability (second row). This prompts us to choose an $\tilde{\alpha}$ that is neither too high nor too low. The selected configurations are listed in rows 2 and 3 of Table~\ref{table:cond_sample_config}.

\begin{table}[t]
\centering
\caption{Best conditional sampling configurations and metrics found via grid search. For reference, the unconditional diffusion samples have variability 1.33 (KS) and 3.67 (NS); reference samples have variability 1.33 (KS) and 4.39 (NS); unconditional diffusion samples have unweighted MELR 0.27 (KS) and 0.37 (NS).}

\vspace{2pt}
{\setlength\tabcolsep{5pt} {\setlength{\extrarowheight}{2pt}
\begin{tabular}{l|c|cccc|}
\cline{2-6}
                                                          & KS        & \multicolumn{4}{c|}{NS}                          \\ \cline{2-6} 
                                                          & $8\times$ & $8\times$ & $16\times$ & $32\times$ & $64\times$ \\ \hline
\multicolumn{1}{|l|}{\% of conditioned elements ($\gamma_{C'}$)} & 12.5      & 1.56      & 0.39       & 0.098      & 0.024      \\
\multicolumn{1}{|l|}{Condition strength ($\tilde{\alpha}$)} & 1.0       & 0.625     & 0.625      & 0.375      & 0.125      \\
\multicolumn{1}{|l|}{SDE steps ($N$)}                       & 256       & 256       & 512        & 1024       & 1024       \\
\multicolumn{1}{|l|}{Sample variability}                    & 0.04      & 0.36      & 1.56       & 3.52       & 3.67       \\
\multicolumn{1}{|l|}{MELR (unweighted)}                     & 0.36      & 0.06      & 0.05       & 0.06       & 0.21       \\ \hline
\end{tabular}
}}
\label{table:cond_sample_config}
\end{table}

\subsection{Downscaling factors}
\label{app:ablation_downscale_factor}

\begin{figure}[t]
\centering
\includegraphics[width=\linewidth]{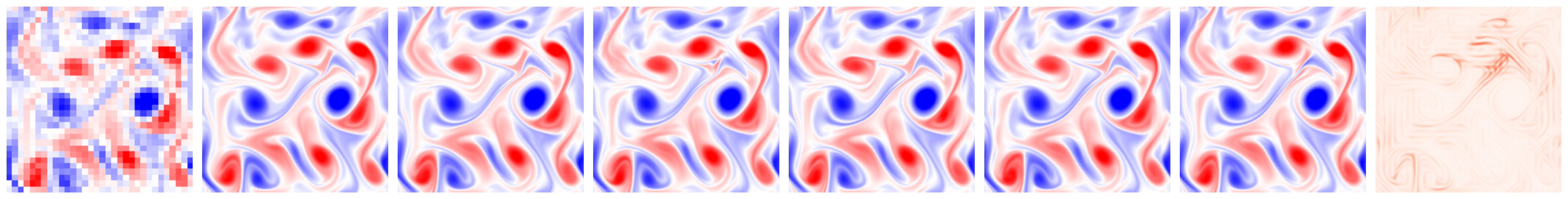}
\includegraphics[width=\linewidth]{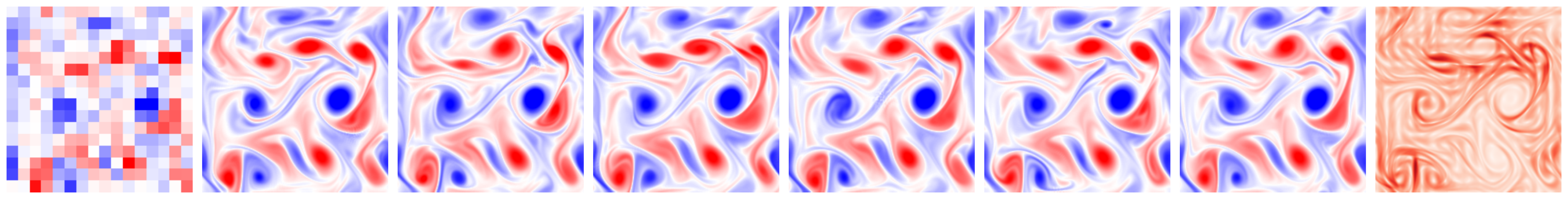}
\includegraphics[width=\linewidth]{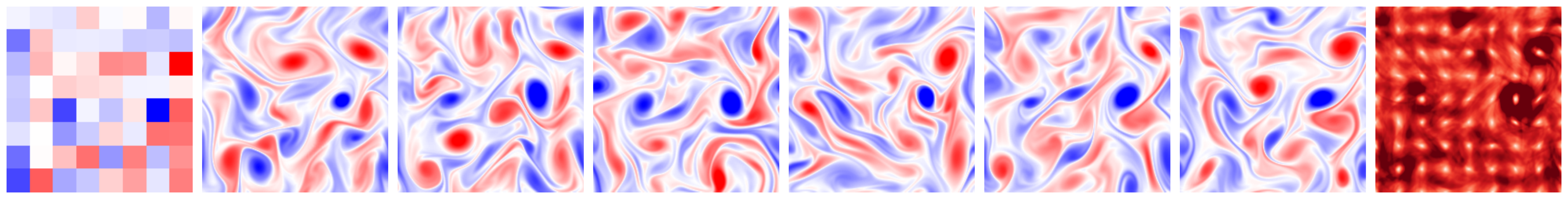}
\includegraphics[width=\linewidth]{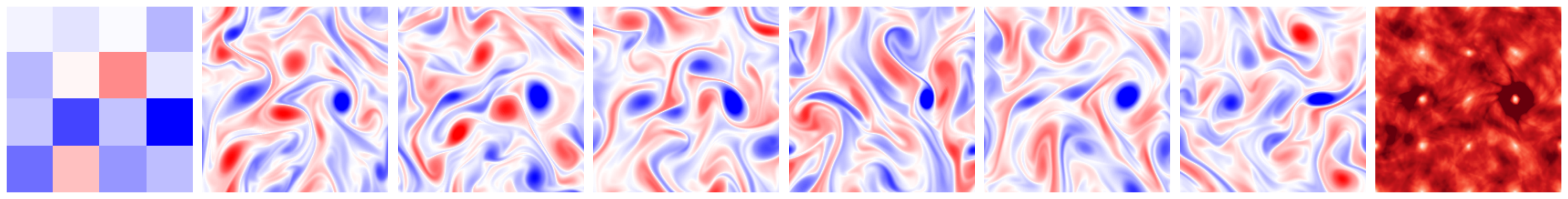}
{\setlength\tabcolsep{0.5pt} {\renewcommand{\arraystretch}{0.5}
  \begin{tabular}{M{.122\linewidth}M{.122\linewidth}M{.122\linewidth}M{.122\linewidth}M{.122\linewidth}M{.122\linewidth}M{.122\linewidth}M{.122\linewidth}}
    (a) & (b) & (c) & (d) & (e) & (f) & (g) & (h) \\
  \end{tabular}
}}
\caption{Sample comparison across different downscaling factors for NS ($8\times, 16\times, 32\times$ and $64\times$ for rows 1, 2, 3 and 4 respectively). Column legend: (a) conditioned values; (b-g) 6 samples generated by diffusion model conditioned on (a); (h) pixel-wise variability of 128 random conditional samples (same color scale across rows; dark means large and light means small).}
\label{fig:ns_cond_samples_with_std}
\end{figure}

We additionally obtain samples for $32\times$ and $64\times$ downscaling (conditioned values are obtained by further downsampling the OT corrected LR snapshots), besides the $8\times$ and $16\times$ presented in the main text, to explore the limits of our methodology. We conduct the same grid search as described in section~\ref{app:ablation_cond_strength} to determine the normalized conditioning strength $\tilde{\alpha}$ and the number of solver steps $N$. The resulting configurations and metrics are displayed in Table~\ref{table:cond_sample_config}, along with samples from an example test case in Fig.~\ref{fig:ns_cond_samples_with_std}. 

We observe that the variability of the generated conditional samples expectedly increases with the downscaling factor, as sampling process becomes less constrained. At $32\times$ downscaling, the corrected LR conditioning still plays a significant role in addressing the color shift in the unconditional sampler, leading to MELR resembling those obtained in the $8\times$ and $16\times$ cases. The same no longer holds true, however, for the $64\times$ downscaling case, as the MELR performance becomes more similar to that of the unconditional sampler. This outcome is also not surprising considering that $64\times$ downscaling corresponds to conditioning on $4\times 4=16$ pixels, accounting for a minuscule 0.02\% of the sample dimensions. 

\subsection{Uncorrected super-resolution}
\label{app:ablation_ot}

To demonstrate the importance of debiasing the low-resolution data, we contrast the performance between conditioning on LR data before and after the OT correction in Table~\ref{table:ns_ot_vs_raw} for all factorized baselines considered. We observe that applying the correction universally leads to better samples regardless of the super-resolution method used.


\begin{table}[t]
\centering
\caption{Metric comparison for conditioning on raw vs. OT-corrected LR snapshots for diffusion-, interpolation- and ViT-based super-resolution.}

\vspace{2pt}
{\setlength\tabcolsep{5pt} {\setlength{\extrarowheight}{2pt}
\begin{tabular}{l|cc|cc|cc|}
\cline{2-7}
                                                       & \multicolumn{2}{c|}{Diffusion}                                & \multicolumn{2}{c|}{Cubic}                                  & \multicolumn{2}{c|}{ViT}                                      \\ \cline{2-7} 
                                                       & Raw                      & OT                                 & Raw                      & OT                                 & Raw                      & OT                                 \\ \hline
\multicolumn{1}{|l|}{MELR (unweighted), NS $8\times$}  & 0.79                     & \textbf{0.06}                      & 0.93                     & \textbf{0.52}                      & 1.39                     & \textbf{0.38}                      \\
\multicolumn{1}{|l|}{MELR (unweighted), NS $16\times$} & 0.54                     & \textbf{0.05}                      & 0.83                     & \textbf{0.55}                      & 1.97                     & \textbf{1.38}                      \\
\multicolumn{1}{|l|}{MELR (weighted), NS $8\times$}    & \multicolumn{1}{l}{0.37} & \multicolumn{1}{l|}{\textbf{0.02}} & \multicolumn{1}{l}{0.41} & \multicolumn{1}{l|}{\textbf{0.06}} & \multicolumn{1}{l}{0.58} & \multicolumn{1}{l|}{\textbf{0.18}} \\
\multicolumn{1}{|l|}{MELR (weighted), NS $16\times$}   & \multicolumn{1}{l}{0.30} & \multicolumn{1}{l|}{\textbf{0.02}} & \multicolumn{1}{l}{0.45} & \multicolumn{1}{l|}{\textbf{0.14}} & \multicolumn{1}{l}{0.32} & \multicolumn{1}{l|}{\textbf{0.10}} \\ \hline
\end{tabular}
}}
\label{table:ns_ot_vs_raw}
\end{table}

\section{Datasets}
\label{app:datasets}

We consider two dynamical systems with chaotic behavior, which is the core property of atmospheric models~\cite{lorenz2005designing}. In particular, we consider the one-dimensional Kuramoto-Sivashinsky (KS) equation and the Navier-Stokes (NS) equation with Kolmogorov forcing. For each equation we implement two different discretizations. The different discretizations are used to generate the low- and high-resolution data. 

\subsection{Equations}

\paragraph{Kuramoto-Sivashinsky (KS) equation} We solve the equation given by 
\begin{equation}
\label{eq:KS_equation}
    \partial _t u + u \partial_x u + \nu \partial_{xx} u  -  \nu \partial_{xxxx} u = 0 \qquad  \text{in } [0, L] \times \mathbb{R}^+, 
\end{equation}
with periodic boundary conditions, and $L=64$.
Here the domain is rescaled in order to balance the diffusion and anti-diffusion components so the solutions are chaotic \citep{Dresdner:2020ml_spectral}. 

The initial conditions are given by 
\begin{equation}
\label{eq:ic_distribution}
    u_0(x) =  \sum_{j = 1}^{n_c} a_j \sin(\omega_j * x + \phi_j),
\end{equation}
where $\omega_j$ is chosen randomly from $\{2\pi/L, 4\pi/L, 6\pi/L\}$, $a_j$ is sampled from a uniform distribution in $[-0.5, 0.5]$, and phase $\phi_j$ follows a uniform distribution in $[0, 2\pi]$. We use $n_c = 30$.

\paragraph{Navider-Stokes (NS) equation} We also consider the Navier-Stokes equation with Kolmogorov forcing given by 
\begin{gather}
\label{eq:NS_equation}
    \frac{\partial \vu}{ \partial t} = - \nabla \cdot (\vu \otimes \vu) + \nu \nabla^2 - \frac{1}{\rho} \nabla p + \mathbf{f} \qquad \text{in } \Omega, \\ 
    \nabla \cdot \vu = 0  \qquad \text{in } \Omega,
\end{gather}
where $\Omega = [0, 2\pi]^2$, $\vu(x,y) = (\vu_x, \vu_y)$ is the field, $\rho$ is the density, $p$ is the pressure, and $\mathbf{f}$ is the forcing term given by 
\begin{equation}
    \mathbf{f} = \left( \begin{array}{c}
        0 \\
        \sin(k_0 y)
        \end{array} 
    \right) + 0.1 \vu,
\end{equation}
where $k_0 = 4$. The forcing only acts in the $y$ coordinate. Following \cite{kochkov_machine_2021}, we add a small drag term to dissipate energy. 
An equivalent problems is given by its vorticity formulation 
\begin{equation}\label{eq:vorticity_NS}
\partial_t\omega = - \vu\cdot\nabla \omega + \nu \nabla^2 \omega  - \alpha\,\omega + f,
\end{equation}
where \mbox{$\omega := \partial_x \vu_y - \partial_y \vu_x$}~\citep{boffetta_two-dimensional_2012}, which we use for spectral method which avoids the need to separately enforce the incompressibility condition~\mbox{$\nabla \cdot \vv = 0$}. The initial conditions are the same as the ones proposed in \cite{kochkov_machine_2021}.

\subsection{Pseudo-Spectral discretization}


To circumvent issues stemming from dispersion errors, we choose a pseudo-spectral discretization, which is known to be dispersion free, due to the \textit{exact} evaluation of the derivatives in Fourier space, while possessing excellent approximation guarantees \citep{trefethen_spectral_2000}. Thus, few discretization points are needed to represent solutions that are smooth. 

We used \texttt{jax-cfd} spectral elements tool box which leverages the Fast Fourier Transform (FFT) \citep{Cooley_Tukey:1965} to compute the Fourier transform in space of the field $u(x, t)$, denoted by $\hat{u}(t)$. Besides the approximation benefits of using this representation, the differentiation in the Fourier domain is a diagonal operator: it can be calculated by element-wise multiplication according to the identity $\partial_x \hat{u}_k = i k \hat{u}_k$, where $k$ is the wavenumber. This makes applying and inverting linear differential operators trivial since they are simply element-wise operations~\citep{trefethen_spectral_2000}.

The nonlinear terms in \eqref{eq:KS_equation} and \eqref{eq:vorticity_NS} are computed using Plancherel's theorem to pivot between real and Fourier space to evaluate these terms in quasilinear time. 
This procedure transforms \eqref{eq:KS_equation} and \eqref{eq:NS_equation} to a system in Fourier domain of the form
\begin{equation}\label{eq:spectral_time_evol}
\partial_t \hat{u}(t) = \mathbf{D} \hat{u}(t) + \mathbf{N}(\hat{u}(t)),
\end{equation}
where $\mathbf{D}$ denotes the linear differential operators in the Fourier domain and is often a diagonal matrix whose entries only depend on the wavenumber $k$ and $\mathbf{N}$ denotes the nonlinear part. We used a 4th order implicit-explicit Crack-Nicolson Runge-Kutta scheme~\citep{Canuto:2007_spectral_methods}, where we treat the linear part implicitly and the nonlinear one explicitly.

\subsection{Finite-volumes discretization} \label{sec:finite_volumes}

We use a simple discretization using finite volumes \cite{leveque_2002}, which was implemented using the finite volume tool-box in \texttt{jac-cfd} \cite{kochkov_machine_2021}.
For the KS equation, we used a Van-Leer scheme to advect the field in time. This was implemented by applying a total variation diminishing (TVD) limiter to the Lax-Wendroff scheme \cite{leveque_2002}. The Laplacian and bi-Laplacian in \eqref{eq:KS_equation} were implemented using tri- and penta-diagonal matrices. The linear systems induced by the implicit step were solved on-the-fly at each iteration using fast-diagonalization.

For the NS equation we used a fractional method, which performs an explicit step which relies in the same Van-Leer scheme for advecting the field together with the diffusion step. We then performed a pressure correction by solving a Poisson equation, also using fast-diagonalization by leveraging the tensor structure of the discretized Laplacian.

\subsection{Data Generation}

For the low-fidelity, low-resolution data (specifically the space $\mathcal{Y}$ in Fig. \ref{fig:framework_diagram}), we employed the finite-volume schemes described above, using either a fractional discretization in time (for NS) or a implicit-explicit method (for the KS equation). The domains mentioned above were utilized, with a $32 \times 32$ grid and a time step of $dt = 0.001$ for NS. For KS, we employed a discretization of size $48$ points and a time step of $dt=0.02$. These resolutions represent the lowest settings that still produced discernible trajectories.

For the high-fidelity, high-resolution data (namely the space $\mathcal{X}$ in Fig. \ref{fig:framework_diagram}), we used the pseudo-spectral discretization mentioned above with a $256 \times 256$ grid and time step $dt = 0.001$ for NS (using the vorticity formulation) and discretization of size $192$ and time step $dt=0.0025$ for the KS equation.  

For the KS equation, we created $512$ trajectories in total. Each trajectory was run for $4025$ units of time, of which we dropped the ones generated during an initial ramp-up time of $25$ units of time. Of the remaining $4000$ units of time, we sampled each trajectory every $12.5$ units of time resulting on $320$ snapshots per trajectory. 

For NS we also created $512$ trajectories in total. We used the same time discretization for both low- and high-resolution data. Each trajectory was run for $1640$ units of time, of which we dropped the ones generated during an initial ramp-up time of $40$ units of time. Of the trajectories spanning the remaining $1600$ units of time, we sampled each them every $4$ units of time (or $4000$ time steps) resulting on $400$ snapshots per trajectory. 

The sampling rate for each trajectory was chosen to minimize the correlation between consecutive snapshots and therefore, obtain a better coverage of the attractor. 

\section{Hyperparameters}
\label{app:hyperparams}

Table~\ref{table:hyper} shows the set of hyperparameters used to train our diffusion models. Our U-Net model (parameterizing $F_\theta$ in \eqref{eq:denoiser_precondition}) closely follows the \emph{Efficient U-Net} architecture in~\cite{saharia2022photorealistic} and apply self-attention operations at the coarsest resolution only. We employ the standard \texttt{adam} optimizer, whose learning rate follows a schedule consisting of a linear ramp-up phase of 1K steps and a cosine decay phase of 990K steps. The maximum learning rate is $10^{-3}$ and the terminal learning rate is $10^{-6}$. We additionally enable gradient clipping (i.e., forcing $\|\text{d}L/\text{d}\theta\|_2\leq1$) during optimization.

\begin{table}[t]
\centering
\caption{Hyperparameters for diffusion model architecture and training.}

\vspace{2pt}
{\setlength{\extrarowheight}{2.5pt}
\begin{tabular}{|l|cc|}
\hline
Hyperparameter                                                               & KS            & NS                    \\ \hline
Input dimensions                                                             & $192\times1$  & $256\times256\times1$ \\
\texttt{Dblock}/\texttt{Ublock} resolutions                                  & $(96, 48, 24)$    & $(128, 64, 32, 16)$          \\
Resolution channels                                                          & $(32, 64, 128)$  & $(32, 64, 128, 256)$         \\
Number of \texttt{ResNetBlocks} per resolution                               & $6$           & $6$                     \\
Noise embedding                                                              & Fourier        & Fourier                 \\
Noise embedding dimension                                                    & $128$         & $128$                   \\
Number of attention heads                                                    & $8$           & $8$                     \\ 
Total number of parameters                                                   & $4.40\text{M}$ & $31.44\text{M}$         \\ \hline
Batch size                                                                   & $512$         & $16$                    \\
Number of training steps                                                     & $1$M          & $1$M                    \\
EMA decay                                                                    & $0.95$        & $0.99$                  \\
Training duration (approximate)                                              & $2$ days      & $4$ days            \\\hline
\end{tabular}
}
\label{table:hyper}
\end{table}

\section{Debiasing with optimal transport}
\label{app:debiasing}

We begin this section by giving an overview of  
computational methods to find optimal transport maps.

For certain measures, the optimal transport plan $\gamma \in \Pi(\mu_Y,\mu_{Y'})$ in the Wasserstein-2 distance $W_2(\mu_Y,\mu_{Y'}) = \inf_{\gamma} \int \frac{1}{2}\|y - y'\|^2d\gamma(y,y')$ 
is induced by a transport map $T \colon \mathcal{Y} \mapsto \mathcal{Y}'$ where $T_\sharp \mu_{Y} = \mu_{Y'}$. In particular for the quadratic cost, Brenier's theorem guarantees that such a map exists when $\mu_{Y'}$ is atom-less~\cite{brenier1991polar} and the plan is concentrated on the graph of a map, i.e., $\gamma(y,y') = (\text{Id}, T)_\sharp \mu_{Y'}$. Moreover, the Brenier map $T$ is given by the gradient of a convex potential function.


Recently, several methods have been proposed to approximate the Brenier map  given only a collection of i.i.d.\thinspace samples from each measure $\{y^i\} \sim \mu_{Y},\{(y')^i\} \sim \mu_{Y'}$. These include flow-based models~\cite{trigila2016data}, the projection arising from an entropic-regularized OT problem as discussed in Section ~\cite{pooladian2021entropic}, 
and continuous approximations of discrete plans~\cite{perrot2016mapping}.  Another recent approach directly parameterizes the transport map as the gradient of a convex potential function that is represented using input convex neural networks~\cite{korotin2019wasserstein, makkuva2020optimal}.
This approach leverages the dual formulation of the OT problem, to express the Wasserstein-2 distance as
\begin{equation} \label{eq:W2_dualproblem}
W_2(p,q)^2 = C_{p,q} + \sup_{f \in \text{cvx}(p)} \left\{\mathbb{E}_p[-f(X)] + \mathbb{E}_q[-f^*(Y)] \right\},
\end{equation}
where $C_{p,q} = \mathbb{E}[X^2] + \mathbb{E}[Y^2]$ is a constant and $f^*(y) = \inf_x \{x^T y - f(x) \}$ is the convex conjugate of $f$. Under the conditions of Brenier's theorem, the optimal map $T$ satisfying $T_\sharp \mu_{Y} = \mu_{Y'}$ corresponds to $T = \nabla f^*$ where $f$ solves~\eqref{eq:W2_dualproblem}.
If we replace $f^*$ with a second network $g$ that is also parameterized with ICNNs,~\cite{makkuva2020optimal} proposed to find the OT map by solving the min-max problem:
$$\sup_{f \in \text{cvx}(p)} \inf_{g \in \text{cvx}(q)} \left\{\mathbb{E}_p[-f(X)] + \mathbb{E}_q[-\langle Y, \nabla g(Y) \rangle - f(\nabla g(Y))] \right\}.$$ This approach  is very sensitive to the network initialization and is challenging to solve in high-dimensions due to the constraints imposed on the map. Moreover, they are limited to squared-Euclidean costs, which limits their flexibility in certain applications. As a result, in our numerical examples we choose to use the entropic OT problem discussed in Section~\ref{sec:OTmaps}.  



\subsection{Additional numerical results}

We provide additional numerical results to showcase how the optimal transport (OT) map corrects the bias in the LFLR snapshots. 

\begin{figure}
\centering
\begin{tabular}{M{.25\linewidth}M{.25\linewidth}M{.25\linewidth}}
\multicolumn{3}{c}{\includegraphics[width=.75\linewidth]{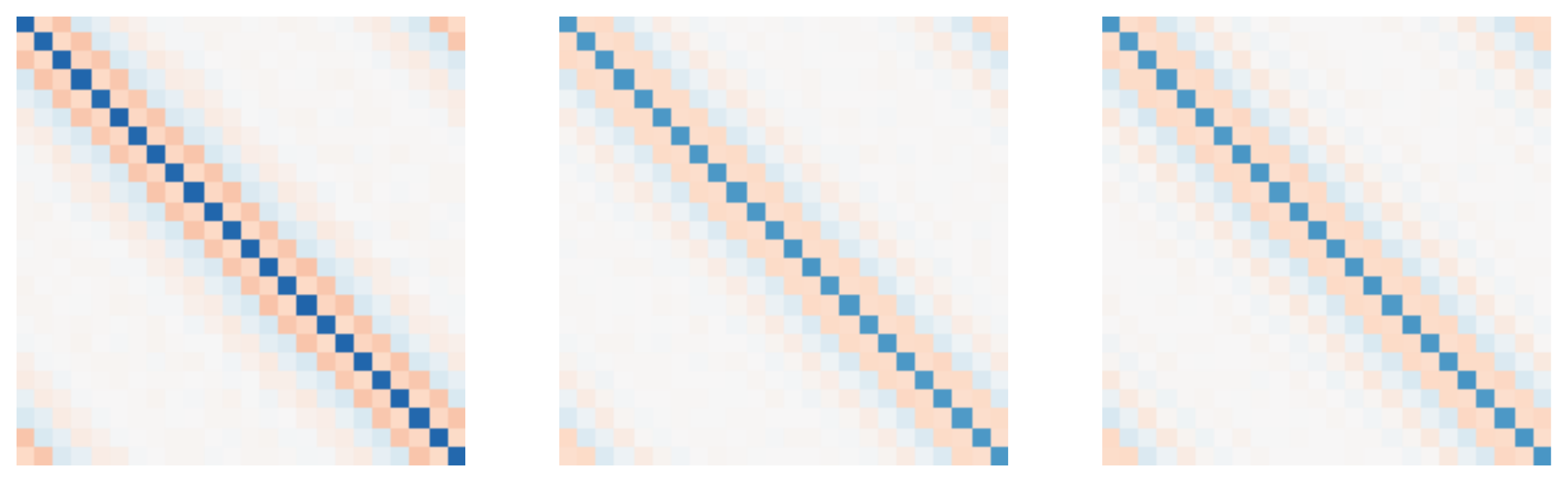}} \\
\multicolumn{3}{c}{\includegraphics[width=.75\linewidth]{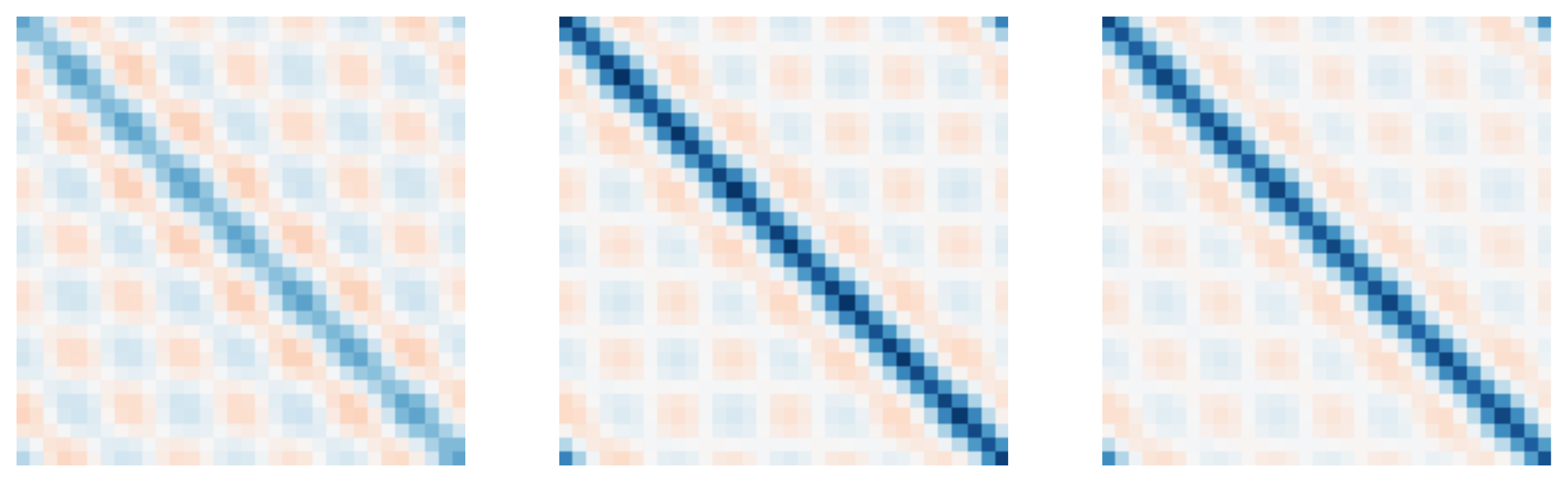}} \\
\multicolumn{3}{c}{\includegraphics[width=.75\linewidth]{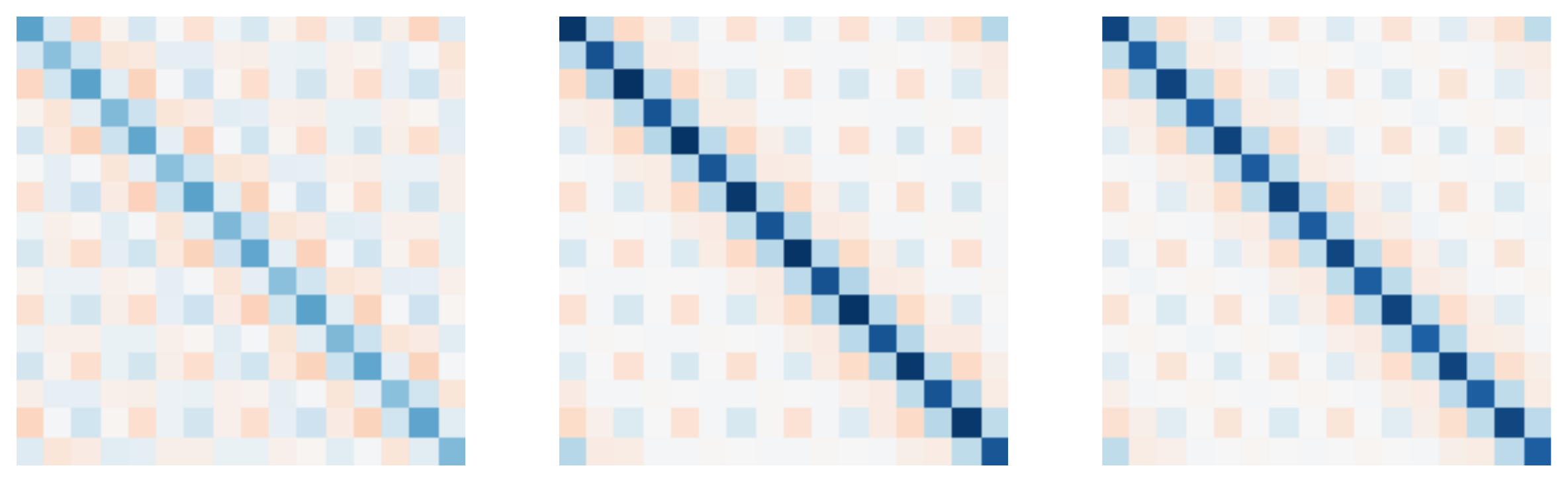}} \\
(a) LFLR   & (b) OT-corrected   & (c) HFLR
\end{tabular}
\caption{Covariance structure of LFLR, OT-corrected and HFLR reference samples for KS (top) NS $8\times$ downscaling (middle) and NS $16\times$ downscaling (bottom).}
\label{fig:covariance_low_res}
\end{figure}

\begin{figure}
\centering
\includegraphics[width=\linewidth]{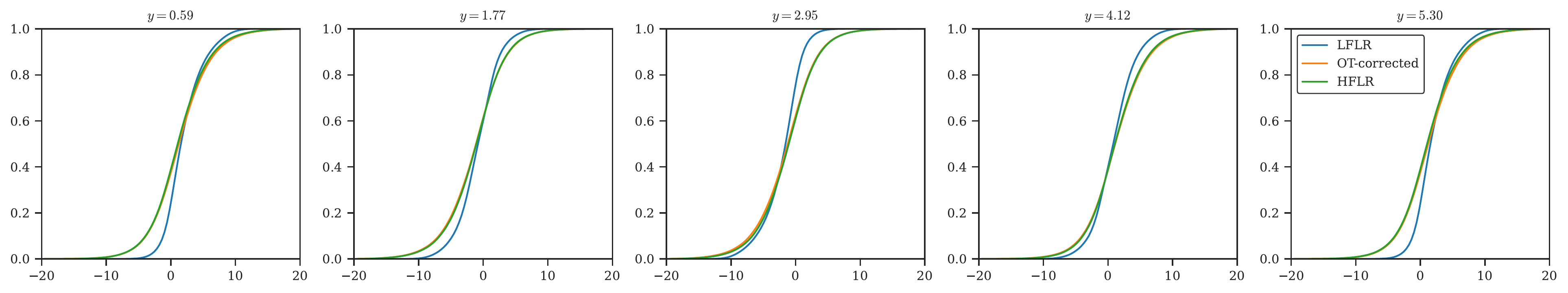}
\includegraphics[width=\linewidth]{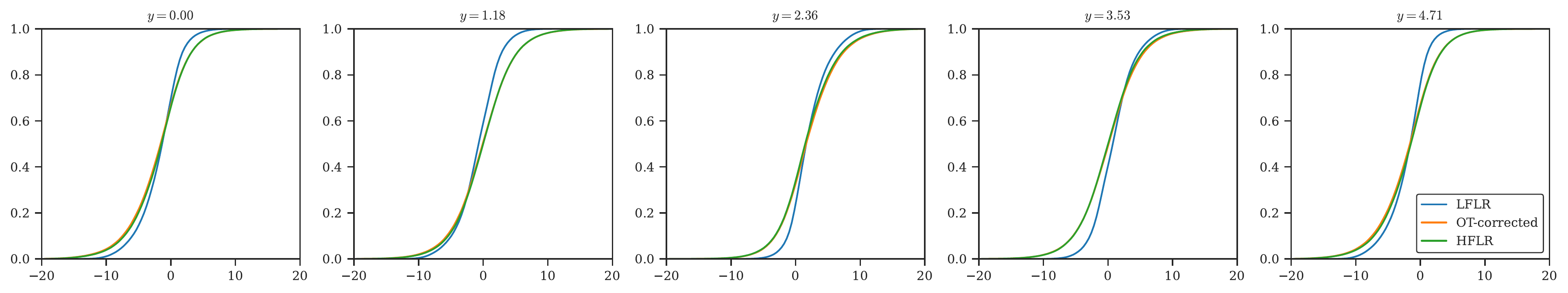}
\caption{Cumulative distribution functions (CDFs) at selected locations of the snapshots for the NS $8\times$ (top) and $16\times$ (bottom) examples.}
\label{fig:ns_ot_cdf}

\end{figure}

Fig.~\ref{fig:covariance_low_res} displays how the OT map changes the covariance structure of the snapshots, while Fig.~\ref{fig:ns_ot_cdf} shows the cumulative distribution functions before and after the OT correction for both the $8\times$ and $16\times$ NS downscaling problems. We can observe from the plots that the OT successfully corrects the distributions.

\section{Computational resources}
The generation of the data was performed using $12$ core server with an NVIDIA A$100$ GPU and $40$ GB of VRAM.
The training for the diffusion models, and the ViT model were performed in a $16$ core server with NVIDIA V100 GPUs with $32$ GB of VRAM. The cycle-GAN was trained on a TPU v4 in Google cloud. The Sinkhorn iteration for computing the OT map was performed in a $80$ core instance with $240$GB of RAM, each training loop took roughly a day for $5000$ iterations. 
All the training was performed in single precision (\texttt{fp32}), while the generation of the data was performed in double precision (\texttt{fp64}). The data was transferred to single precision at training/inference time. 

\section{Additional samples}

\begin{figure}[p]
\centering
\includegraphics[width=\linewidth]{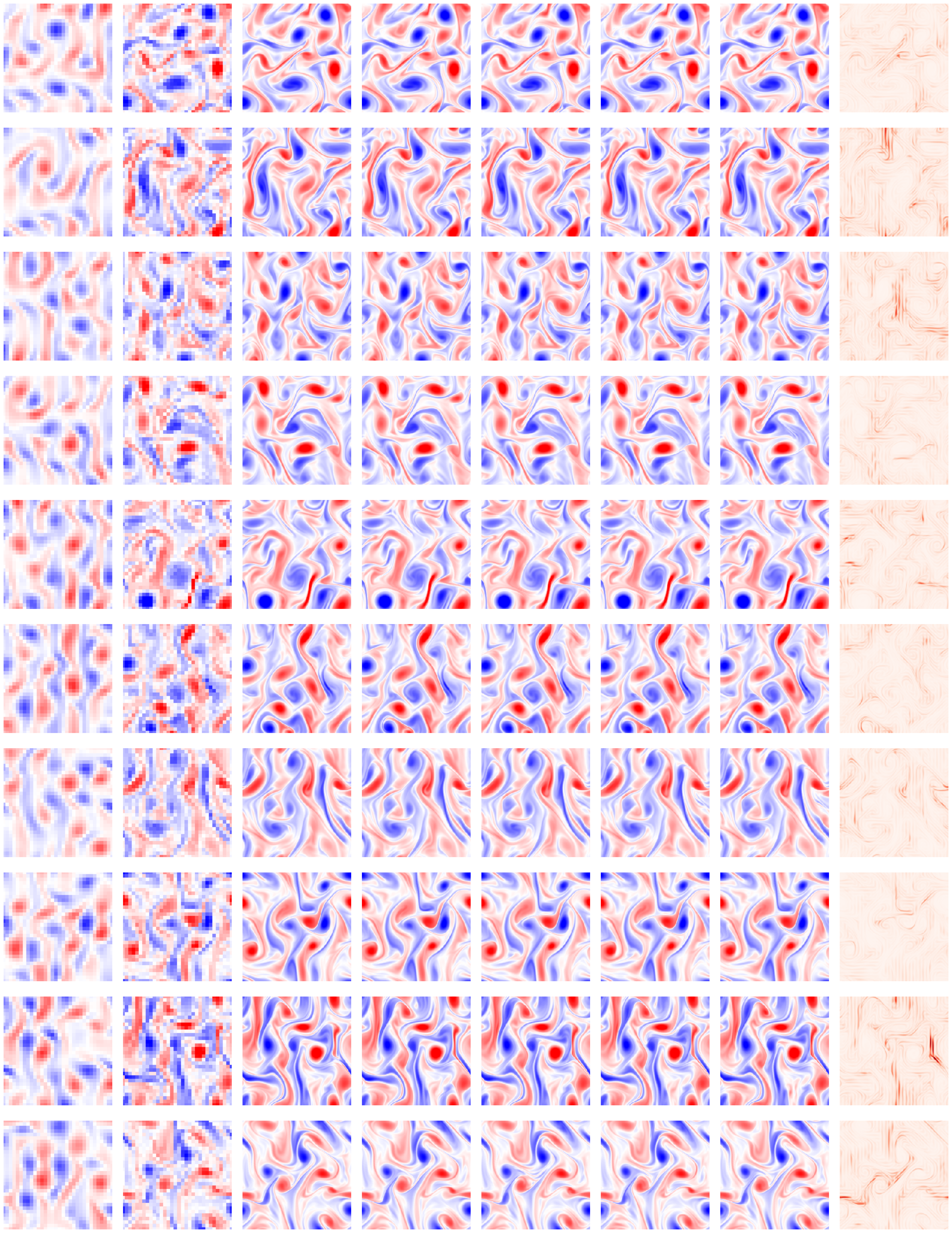}
{\setlength\tabcolsep{0.5pt} {\renewcommand{\arraystretch}{0.5}
  \begin{tabular}{M{.122\linewidth}M{.122\linewidth}M{.122\linewidth}M{.122\linewidth}M{.122\linewidth}M{.122\linewidth}M{.122\linewidth}M{.122\linewidth}}
    (a) & (b) & (c) & (d) & (e) & (f) & (g) & (h) \\
  \end{tabular}
}}
\caption{Conditional samples for NS 8$\times$ downscaling. Column legend: (a) raw LFLR snapshot; (b) LFLR snapshot corrected by OT; (c-g) 5 samples generated by diffusion model conditioned on (b); (h) pixel-wise variability of 128 random samples conditioned on (b).}
\label{fig:extra_NS_8x}
\end{figure}

\begin{figure}[p]
\centering
\includegraphics[width=\linewidth]{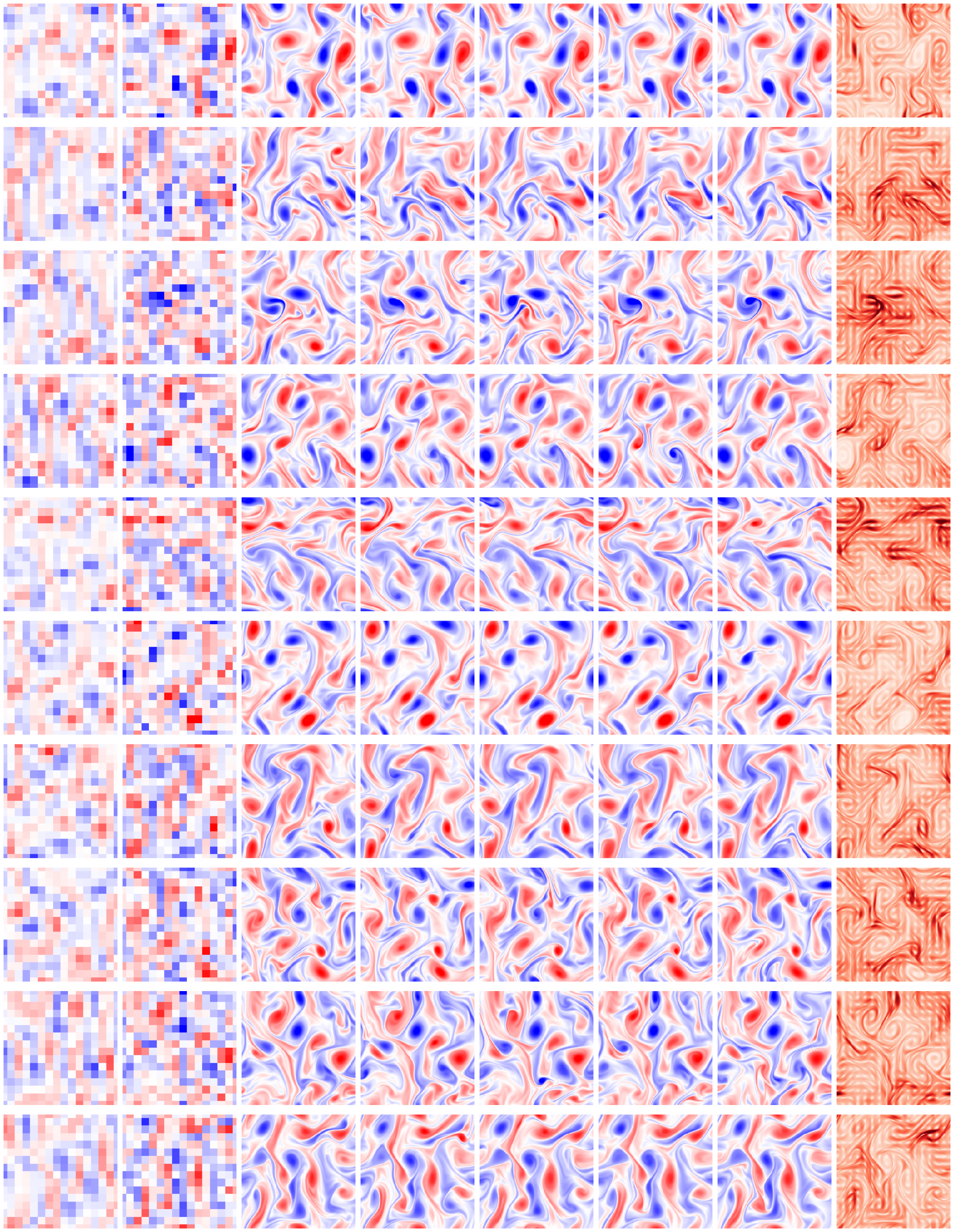}
{\setlength\tabcolsep{0.5pt} {\renewcommand{\arraystretch}{0.5}
  \begin{tabular}{M{.122\linewidth}M{.122\linewidth}M{.122\linewidth}M{.122\linewidth}M{.122\linewidth}M{.122\linewidth}M{.122\linewidth}M{.122\linewidth}}
    (a) & (b) & (c) & (d) & (e) & (f) & (g) & (h) \\
  \end{tabular}
}}
\caption{Conditional samples for NS 16$\times$ downscaling.  Column legend: (a) raw LFLR snapshot; (b) LFLR snapshot corrected by OT; (c-g) 5 samples generated by diffusion model conditioned on (b); (h) pixel-wise variability of 128 random samples conditioned on (b).}
\label{fig:extra_NS_16x}
\end{figure}

We provide additional conditional samples in Figs.~\ref{fig:extra_NS_8x} and~\ref{fig:extra_NS_16x} from the NS $8\times$ and $16\times$ downscaling experiments respectively.

\end{document}